\documentclass[review]{elsarticle}

\usepackage{hyperref}
\usepackage{amsmath}
\usepackage{multirow}
\usepackage{graphicx}
\usepackage{lscape}
\usepackage{adjustbox}
\usepackage{xcolor}
\usepackage{soul}
\usepackage{booktabs}  
\usepackage{amsmath}
\usepackage{textcomp}  
\usepackage{lipsum}    
\usepackage{subfig}    
\usepackage{hyperref}
\usepackage{multirow}
\usepackage{graphicx}
\usepackage{lscape}
\usepackage{adjustbox}
\usepackage{pdflscape}
\usepackage{everypage}
\usepackage{lipsum}

\newcommand{\Lpagenumber}{\ifdim\textwidth=\linewidth\else\bgroup
  \dimendef\margin=0 
  \ifodd\value{page}\margin=\oddsidemargin
  \else\margin=\evensidemargin
  \fi
  \raisebox{\dimexpr -\topmargin-\headheight-\headsep-0.5\linewidth}[0pt][0pt]{%
    \rlap{\hspace{\dimexpr \margin+\textheight+\footskip}%
    \llap{\rotatebox{90}{\thepage}}}}%
\egroup\fi}
\AddEverypageHook{\Lpagenumber}

\journal{Expert Systems with Applications}

\biboptions{authoryear}

\begin{document}

\begin{frontmatter}

\title{Loss Rate Forecasting Framework Based on Macroeconomic Changes: Application to US Credit Card Industry}

\author[mymainaddress]{Sajjad Taghiyeh\corref{mycorrespondingauthor}}
\cortext[mycorrespondingauthor]{Corresponding author}
\ead{staghiy@ncsu.edu}
\author[mysecondaryaddress]{David C Lengacher}
\author[mysecondaryaddress2]{Robert B Handfield}

\address[mymainaddress]{North Carolina State University, Raleigh, NC USA \\ email: \href{mailto:staghiy@ncsu.edu}{staghiy@ncsu.edu} }
\address[mysecondaryaddress]{Raynolds American, Winston-Salem, NC USA \\ email: \href{mailto:lengacd@rjrt.com}{lengacd@rjrt.com}}
\address[mysecondaryaddress2]{North Carolina State University, Raleigh, NC USA \\ email: \href{mailto:rbhandfi@ncsu.edu}{rbhandfi@ncsu.edu}}

\begin{abstract}
A major part of the balance sheets of the largest US banks consists of credit card portfolios. Hence, managing the charge-off rates is a vital task for the profitability of the credit card industry. Different macroeconomic conditions affect individuals' behavior in paying down their debts. In this paper, we propose an expert system for loss forecasting in credit card industry using macroeconomic indicators. We select the indicators based on a thorough review of the literature and experts' opinions covering all aspects of the economy, consumer, business, and government sectors. The state of the art machine learning models are used to develop the proposed expert system framework. 

We develop two versions of the forecasting expert system, which utilize different approaches to select between the lags added to each indicator. Among 19 macroeconomic indicators that were used as the input, six were used in the model with optimal lags, and seven indicators were selected by the model using all lags. The features that were selected by each of these models covered all three sectors of the economy. Using the charge-off data for the top 100 US banks ranked by assets from the first quarter of 1985 to the second quarter of 2019, we achieve mean squared error values of 1.15E-03 and 1.04E-03 using the model with optimal lags and the model with all lags, respectively. The proposed expert system gives a holistic view of the economy to the practitioners in the credit card industry and helps them to see the impact of different macroeconomic conditions on their future loss.

\end{abstract}


\end{frontmatter}


\section{Introduction}\label{intro} 


Similar to any industry, the goal in the consumer credit industry is to maximize profits by measuring and controlling risk and avoiding exposure to default (also known as charge-off), as much as possible. The term charge-off means an outstanding credit card debt, which is written off as bad debt. Consumers must issue payments by the due date, and failure to do so will result in putting the consumer's account into delinquency or default. Typically, a bad credit card debt will be marked as charged-off after six months of non-payment, and it is withdrawn as an asset from the lender's accounts. This is usually a final action since it is an indication to lenders that the consumer will never pay off their account. Thus the account is written-off as bad debt. The charge-off rate for a given bank or issuer is calculated by dividing the dollar amount of charge-offs by average outstanding balances on credit cards issued by the firm. A higher charge-off rate exhibits a higher risk to a company. Usually, strategic business analysis is incorporated by credit card issuers to develop credit policy and guidelines with legal and regulatory constraints. Credit policy helps an institution develop strategies within the planned asset quality range that are consistent with the institution's profitability goals. Accurate prediction of charge-off rates has been one of the major challenging tasks in the credit card industry. The charge-off rate has shown a strong tie to economic conditions, and it has hit its highest level during the financial crisis, which was 10.79\% according to U.S. Federal Reserve data. Increasing the charge-off rates during the 2008 financial crisis led to the question of how we can predict the charge-off rate based on macroeconomic indicators under different economic conditions.

There has been extensive research on the relationship between charge-off risk and general economic climate, resulting in a general belief that macroeconomic factors directly affect bad debts and charge-offs. Historical data obtained from credit bureaus along with consumer performance data are analyzed by lenders to predict the future behavior of consumers and their risk of going delinquent or charging off. These predictive models classify consumers into different segments and align the bank's strategies towards these segments accordingly. The problem is that many businesses rely only on these models to make decisions, and fail to include certain economic factors into their risk models. Sometimes, to include economic conditions, these predictive models are adjusted by several percentage points in the charge-off rate using a fraction of macroeconomic indicators. However, most of the time, only a fraction of economic aspects are reviewed for these adjustments, as they are deemed to be the most influential.

Consumers' charge-off behavior can be heavily affected as the economy goes through good times (expansion phase) and bad times (the contraction phase), and they are not explicitly modeled in prediction models developed by credit risk management, which raises the question of how charge-off rate will change in different economic conditions. During economic expansion, consumers and businesses have enough income to pay their debts by their respective due dates, and thus this phase is associated with a small number of delinquencies and charge-offs. On the other hand, in the contraction phase, the number of bad debts will increase, which eventually will lead to a significant jump in the charge-off rate. Credit card companies can be affected by economic factors, and including economic factors in the decision-making process may significantly impact their ability to make effective charge-off decisions proactively. Failing to incorporate economic factors may lead to consequences that may take years for the company to recover. Since many other factors such as government regulations are already reducing the profits of credit card business, there is a need for a new approach that incorporates the relationship between economic factors and charge-off.

Early credit card portfolio literature could not find conclusive evidence on the effects of macroeconomic factors on charge-offs over the business cycle. For example, personal bankruptcy and credit card delinquencies in the 1990s were investigated in  \citet{gross2002empirical}, and authors concluded that the relationship between charge-offs and macroeconomic factors had changed substantially over the investigation period and there was not conclusive evidence to prove a relationship between charge-off rate and macroeconomic factors. They also concluded that the unemployment rate has no significant impact on the charge-off rate. They used panel data on credit card accounts for their analysis. However, later in \citet{agarwal2003determinants}, the authors stated that the unemployment rate has significant predictive power for the charge-off rate. They noted that the reason behind the fact that previous empirical studies could not find a consistent relationship between economic factors and bankruptcy is that those studies were either suffering from inadequate data or the variation in the unemployment rate was not sufficient during their analysis period. 

Following the Great Recession in 2008, credit card companies focused heavily on controlling credit losses. Their emphasis is mostly on the unemployment rate, as it has a strong correlation with the charge-off rate. However, in the past few years, the unemployment rate was going down while the charge-off rate was increasing, and a model using unemployment rate as its only input may not be able to capture the uptrend in the charge-off rate. Hence, credit card companies need to focus on other economic factors that can affect charge-offs, and most importantly, they need to look at the economy as a whole. Analyzing the impact of variables from all segments of the economy will provide lenders with a holistic insight and will help them to make more effective decisions to reduce future losses. 

There are limited cases in the body of literature that focus on charge-off prediction models incorporating macroeconomic variables in the United States. The slope of U.S. Treasury bond yields over time was mentioned by \citet{estrella1991term} and \citet{estrella1998predicting} to have a strong relationship with output growth and recessions in the United States up to eight quarters in the future. Stock prices \citep{estrella1998predicting}, credit market activity \citep{levanon2011using}, index of leading economic indicator \citep{berge2011evaluating,stock2002forecasting} and several interest rates, housing indices and unemployment rate measures \citep{ng2014boosting} as leading indicators for future economic conditions. Moreover, there are different views regarding the significance of specific economic factors. For instance, industrial production was found to be a significant predictor of corporate charge-offs by \citet{figlewski2012modeling}. However, research was done by \citet{giesecke2011corporate} has shown that it may not be an important factor in forecasting the charge-off rate. Stochastic optimization algorithms can also be used in financial industry to improve the efficiency of the algorithms \citep{taghiyeh2016new}.

The author was motivated to perform this study when he started working as analytics intern at one of the leading credit card issuer companies in the United States. The models in production were using only unemployment rate as their input to forecast future values of the loss rate, which had an R-squared value of about 63\%. Aside from the relatively low R-squared value, the charge-off rate was going up in the past couple of years, but unemployment rate was going down. Therefore, their model was unable to predict uptrend in charge-off rate and it was crucial to develop a new prediction model for the charge-off rate by incorporating macroeconomic factors from all aspects of the economy.

In this study, we aim to identify and analyze economic indicators that have a significant relationship with the charge-off rate in the credit card industry. Next, we will use machine learning techniques, namely, linear regression with Lasso, linear regression with Ridge, random forest, and gradient boosting machine to develop a loss forecasting framework using selected macroeconomic indicators. Finally, using the model selection approach introduced in \cite{taghiyeh2020forecasting} (MSIC algorithm), we will forecast each of the selected indicators to predict year over year changes. Nineteen macroeconomic indicators from three major economic categories will be used for this analysis. These economic categories include consumer, business, and government segments. The use of indicators from all segments gives a comprehensive view of the economic impact on charge-offs. Credit card companies have recently identified the unemployment rate and housing indices as charge-off accelerators. These two metrics will be included in our analysis to confirm or deny their assumptions. The consumer confidence index is another factor that can be seen to have an impact on charge-off rates, as consumer behavior may change payment behavior when they are optimistic or pessimistic towards the future. However, this index is very volatile and may fluctuate each month as the report comes out \citep{censky2010consumer}. Other macroeconomic indicators used in this research are new from a charge-off analysis standpoint. Charge-off data from the top 100 banks in the United States from 1985 to 2019 will be used in this study to confirm if the selected macroeconomic indicators have significant predictive power for the duration of the analysis. The design of a prediction model covering all aspects of the economy will add a significant value to a company. Executives and managers can incorporate this information into their decision process to anticipate any future credit losses and fluctuations.

The remainder of this paper is organized as follows. Section 2 reviews the literature on loss forecasting. Section 3 presents the details of our proposed loss forecasting framework based on macroeconomic indicators. Section 4 presents an empirical evaluation of the approach using loss data from the top 100 banks in the United States. We summarize our conclusions and discuss the practical implications of our work in section 5.

\section{Literature Review}\label{litRev}
Credit card companies are in the business of lending money to consumers, but it is a very risky task as they are not certain if consumers will pay back their debt or make payments by due dates \citep{guseva2001uncertainty}. To assess the probability of charge-off, credit card issuers usually use scoring models, which are mostly based on historical consumer performance gathered from any of the credit bureaus, such as Equifax, Experian, and Transunion. This information is used to develop account level models to evaluate the risk of a particular consumer and divide them into low-risk and high-risk segments. However, the account level model may put consumers in the low-risk segment, but eventually, they do not pay their debt and charge-off. What would have been the reason behind the fact that the low-risk consumers ended up being charge-off or delinquent? What are the non-credit factors that affected the payment ability of the consumers that were not taken into account in account level risk models? The answer to these questions may be uncertainty derived from underlying economic conditions, which need to be taken into account for by credit card companies when offering credit to consumers.

The main issue for credit card companies is when to take action and to what extent they need to tighten their credit offerings. If the right time is chosen to act, it may lead to a stable or even increasing revenue, and lenders will avoid unnecessary charge-offs and loss. Nevertheless, an ill-timed action will bring the company an increased loss and a steep shrinkage in revenue. Hence, credit card companies face credit loss challenges brought upon by strong and weak economic conditions. During the 1990s, people kept spending using credits and generated high balances on their credit cards, which lead to a significant loss due to non-payment on debts when the recession hit in 2001 \citep{evans2005paying}. It was the same situation for many credit companies during the 2008 great recession. One of the significant issues with recessions is an increase in the unemployment rate. This will greatly affect the ability of consumers to pay their unpaid debt and credit card charge-offs. If the charge-off trend is identified early in the phase, it will give credit card companies enough time to make the right decision and act promptly and avoid unnecessary losses or drop in revenue. Evaluating the impact of the economy on the charge-off rate would help lenders to predict the trend and make effective decisions.

In the last recession, several banks started to tighten their credit offering criteria in the last quarter of 2007, but an aggressive action was not taken until the second half of 2008, in which the unemployment rate was already risen by 30\%. Credit card issuers tightened the credit offerings by closing the accounts and reducing credit lines. However, the charge-off rate has hit its highest at 10\% in the last quarter of 2009, and the credit card company's revenue had a steep shrinkage. Also, Many good customers were affected by this sudden reaction. At that time, it became clear that the unemployment rate and charge-off rate have a strong correlation, and many banks today continue to use the unemployment rate as one of the decision factors in their strategies. However, there are many other economic indicators that may help lenders to understand the future of the economy better and predict the charge-off trend.

To the best of our knowledge, there is only one research that studies the relationship between economic factors and the charge-off rate in the US economy \citep{liu2003predictive}. In the empirical study by \citet{liu2003predictive}, authors use step-wise regression and vector autoregression to identify economic factors which have predictive power regarding credit card charge-offs. The goal of their research was to develop a predictive model based on these variables. Authors concluded that the unemployment rate, consumer confidence index, household debt service burden, inflation rate, personal bankruptcy filings, and stock market returns are the variables that have a strong predictive power for the charge-off rate. However, there are a few issues with their work that justifies a more recent and thorough work toward identifying economic variables to develop a predictive model for charge-off. The first issue is that their analysis is focused on the period of 1986-1998, and there were quite a few changes in both the credit card industry and economic conditions. Second, in \citet{liu2003predictive}, authors only include seven economic variables in their analysis, and they are not covering all the aspects of the economy, namely government, business, and households, entirely.

There exist several studies on the relationship between charge-off and economic conditions. It was stated by \citet{ausubel1997credit} that in a generally healthy economy, in which unemployment is relatively low and gross domestic product is reasonably growing, both bankruptcy and charge-off rate increased. This statement was against the general belief that the charge-off rate will increase during bad economic times and decrease in good economic times, and has shown that other economic factors may contribute to charge-offs. Debt-to-disposable income ratio was found by \citet{stavins2000credit} to have a strong correlation with credit card charge-off and bankruptcy. The relationship between consumer debt burden and economic indicators was studied by \citet{schmitt2000does}, and it was concluded that consumers' debt burden increases in the expansion phase of the economy. The author found personal consumption expenditure, durable goods, and retail sales to have the most predictive power toward consumer debt and the installment loan delinquency rate. 

Since credit cards provide a more flexible way comparing to installment loans, the variables mentioned above may have a different impact on credit card charge-offs. The unemployment rate, consumer price index and the number of bankruptcy filings were deemed to be highly correlated with the charge-off rate in the case of Hong Kong \citep{fung2002modeling}. The authors used the vector regression model as the basis for their analysis. Macroeconomic indicators were analyzed by \citet{agarwal2003determinants} to investigate credit card delinquency. The authors conclude that macroeconomic fluctuations correlate with bankruptcy and delinquency rates. They also found that the unemployment rate has a strong effect on the rate of delinquency. In the analysis done by \citet{musto2006portfolio}, the covariance of individual charge-off risk with aggregate charge-off rate was calculated, and it was found that a significant heterogeneity in the covariance of risk exists among consumers. They also stated that the credit line decreases when the covariance of risk increases. By applying portfolio theory to consumer lending, \citet{desai2014county} extend the work of \citet{musto2006portfolio}. Authors use credit score along with charge-off and bankruptcy rates to evaluate the charge-off. \citet{mian2011house} investigate the relationship between household borrowing and house prices by analyzing account level datasets. The authors conclude that there is a significant relationship between these two variables, and housing prices and household debts can explain fluctuations in the economy. These authors also state that the effect of fluctuations in housing prices is not homogeneous across the population, and people with low credit scores, which highly leverage their credit, are more sensitive to these changes in housing prices. 

A regime-switching model is used by \citet{giesecke2011corporate} to evaluate the predictive power of macroeconomic variables for charge-off rates. Changes in the gross domestic product, stock returns, and their volatility were identified as significant variables. Reduced-form Cox intensity models were fit by \citet{figlewski2012modeling} to analyze the relationship between a range of macroeconomic and firm-specific factors and charge-off and significant credit ratings. They found that both factor categories were significant, but macroeconomic variables were highly dependent on the inclusion of other factors. Using account-level data, \citet{bellotti2012loss} compare the performance of several loss forecasting models, including a decision tree and fractional logit transformation. The authors conclude that using macroeconomic variables in ordinary least square models will result in the best forecasting model. In an extension to their work, in \citet{bellotti2013forecasting}, a discrete time survival model was proposed to predict the probability of charge-off. They claim that using macroeconomic variables along with behavioral factors, the best fit will be obtained. Borrowers' characteristic was also found in \citet{leow2014intensity} to impact charge-off and recovery behavior significantly. In the study of \citet{rubaszek2014determinants}, interest rate spread and income uncertainty were found to impact the amount of household credit using both theoretical and empirical models. 

To evaluate the effect of FICO score, debt-to-income ratio, credit grade, and credit utilization on charge-off, logistic regression, and Cox proportional hazard models were used by \citet{emekter2015evaluating}. The authors concluded that the probability of charge-off increases as debt age increases. A classification model based on the random forest for good and bad loans was proposed by \citet{malekipirbazari2015risk}. The results were compared to the ones obtained from logistic regression, support vector machine, and K-nearest neighbor. However, none of the last two studies we mentioned used out of sample performance to evaluate their classification models. In a study performed by \citet{guo2016instance}, cross-validation was used to evaluate out of sample performance for the credit assessment model of P2P loans. The relationship between the age of the borrower and the probability of charge-off in the US was investigated by \citet{debbaut2016card}. The authors conclude that the probability of charge-off is lower in younger borrowers. Using macroeconomic indicators in Turkey, \citet{mazibacs2017understanding} analyze the reason behind recent fluctuations in household debt. In a study performed in Korea, \citet{kim2017additional} use account-level credit data to show a positive relationship between the probability of delinquency and the amount of debt. A model with bankruptcy, delinquency, and renegotiation was proposed by \citet{kovrijnykh2017screening}. Authors conclude that instead of taking charge-off as a binary event, one needs to look at it as a multiple-stage process.

As the literature review performed in this section suggests, the basis of this research is supported by scholars in the field. As we can see, most of the researches believe that macroeconomic factors affect lenders and financial institutions, and by studying the effects of macroeconomic indicators, we can have a better perception of future lending risks. It is essential for credit card companies to incorporate macroeconomic indicators in their risk models to predict future risks and operate effectively in both the expansion and contraction phases of the economy. This way, they can avoid any unnecessary loss in their portfolio due to a lack of perspective toward economic conditions. Several economic factors were studied in previous researches regarding the charge-off rate. However, in this study, we will cover more economic indicators that encompass all segments of the economy, namely households, government, and business segments. Credit card issuers suffer from unexpected charge-offs due to lack of insight from economic conditions, and a charge-off prediction model which is based on macro-economy data will help managers to make effective business and strategic decisions. This research aims to fill this gap and find the economic indicators with the most significant power to predict future charge-off rates and will use these indicators to build a loss forecasting model for predicting charge-off rate using machine learning models.

\section{Methodology}
In this section, we will use machine learning tools to develop the loss forecasting framework. First, we will explain the important trade-off between interpretabiliy and accuracy that is a hot topic when it comes to using machine learning models, and we will discuss the reason behind the selection of machine learning models in our proposed loss forecasting framework.

Based on the literature review, several macroeconomic indicators that were likely to have correlations with the charge-off rate were selected. Among the selected macroeconomic indicators, 19 indicators were selected by the experts in the credit card industry to form the basis of this research. The goal is to use these indicators as independent variables in a machine learning based model to predict the charge-off rate, which is our dependent variable. In the first step, we apply different transformations (e.g., square root, exponential, ...) to normalize the selected indicators and find the transformation with the highest correlation to dependent variables. We also add lags from 1 to 4 quarters to each indicator and find the correlation of each of the lagged indicators with charge-off rate. This way, we incorporate the lagged effects of each macroeconomic indicator. The next step in data preparation is to convert all indicators and charge-off rate to year over year changes. To do so, for each indicator, we record the percentage of change comparing to the corresponding period in the last year. This way, instead of using the actual values for macroeconomic indicators to predict the charge-off rate, we build a model that uses the changes in each indicator to forecast the change in charge-off rate.

After we have generated our input data, we will use two versions of Lasso regression (Lasso with optimal lags and Lasso with all lags) to select the features with the most significant correlation to our output data. The difference between these two feature selection methods lies in the approach we use to generate their input. In the first feature selection model (Lasso with optimal lags), for each indicator, we select the lag, which has the highest correlation with the charge-off rate. Therefore, the model has 19 independent variables corresponding to optimal lags for each of the selected macroeconomic indicators. In the second approach, which is Lasso with all lags, we include all the lags in the input data and let the model select between lags. Note that, in the second feature selection method, we let the model choose more than one lag from each indicator. In doing so, the model can capture the trends for the year-over-year changes of each macroeconomic indicator. 

We use the indicators selected by each of the feature selection methods as the input to train machine learning models and capture the relationship between the selected macroeconomic indicators and the charge-off rate. As mentioned earlier, the benchmark machine learning models in this study are Lasso regression, Ridge regression, gradient boosting machine (GBM), and random forest (RF). As it is common in the machine learning field, we split the data into training and test sets to train and evaluate the performance of each machine learning model. Two sets of machine learning models need to be developed since we have two versions of input data resulted from different feature selection approaches.

\begin{figure}[htbp]
        \center{\includegraphics[width=0.7\textwidth]
        {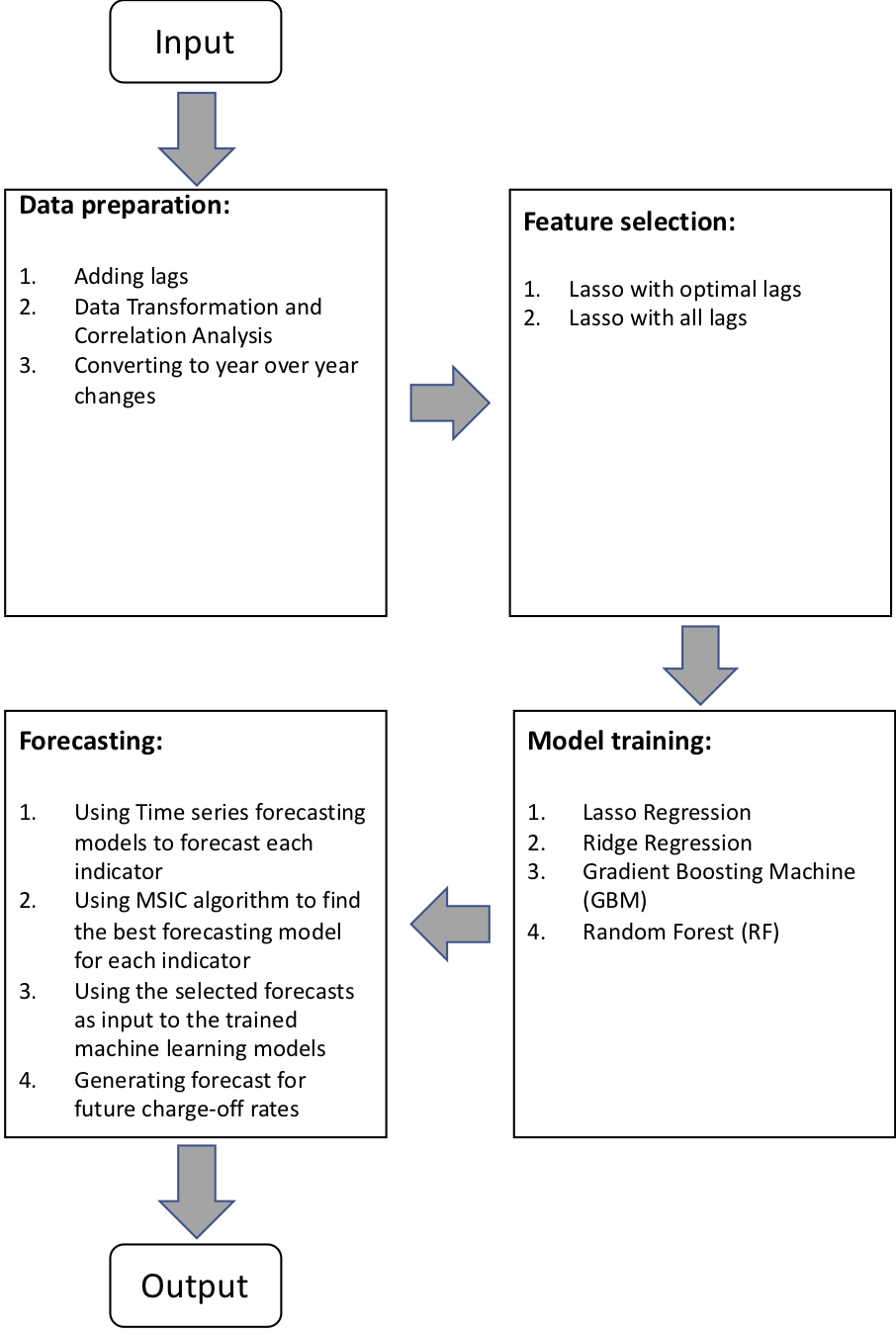}}
        \caption{\label{Loss_Forecasting} Steps to develop the proposed loss forecasting framework }
      \end{figure}

The last piece of building the loss forecasting framework is to predict future values for each of the selected macroeconomic indicators and use the trained machine learning model to predict future charge-off levels. To predict each macroeconomic indicator, seven well-known forecasting models have been used, namely, nai\"ve forecasting, moving average, simple exponential smoothing, Holt, Holt-Winters, ARIMA, and Theta. These models are selected among the models considered in the forecasting competitions, such as M3-Competition. Three variants of the MSIC algorithm proposed in \cite{taghiyeh2020forecasting} are used to select the best performing forecasting model for each macroeconomic indicator. Using the results from the forecasting model selected by the MSIC algorithm, the trained machine learning models are then used to predict the future values of the charge-off rate. Figure \ref{Loss_Forecasting} shows the steps of the loss forecasting framework proposed in this study.  The details of our proposed Loss rate forecasting framework is outlined in the following subsection.

\subsection{Loss Forecasting Algorithm}
Let n be the number of macroeconomic indicators that are selected to build the loss forecasting framework and let m be the number of machine learning models used to predict the loss rate. Feature selection method can be set to either "Optimal lags" or "All lags".
\begin{itemize}
    \item Step 1 (data preparation): Initialize final input values as $I_1=\{\}$. For each macroeconomic indicator $i$ $(i=1, ..., n)$:
    \begin{itemize}
        \item Step 1-1: Convert macroeconomic indicator $i$ into quarterly values.
        \item Step 1-2: Add lags from 1 to 4 quarters to indicator i and record the lagged indicator.
        \item Step 1-3: Try different transformations (e.g., square root, exponential, square, log, etc.) for each lagged indicator and select the best one based on a goodness of fit statistic.
        \item Step 1-4: For each lagged indicator, add the selected transformation in step 1-3 to $I_1$.
        \item Step 1-5: Convert all the lagged indicators to year over year values by dividing them by the corresponding values from last year.
        \item Step 1-6: If $i=n$, go to step 2. Else, set $i=i+1$ and go to step 1-1.
    \end{itemize}
    \item Step 2 (feature selection): If "Optimal lags" is selected for feature selection, go to step 2-1. Else, if "All lags" is selected go to step 2-2.
    \begin{itemize}
        \item Step 2-1 (feature selection with optimal lags): Initialize the input data for feature selection as $I_2=\{\}$ and list of final selected features as $F$. Use loss rate as dependent variable.
        \begin{itemize}
            \item Step 2-1-1: For each macroeconomic indicator $i$ $(i=1, ..., n)$, select the lag with the highest correlation with loss rate from $I_1$ and append it to $I_2$.
            \item Step 2-1-2: Apply Lasso regression using the input data $I_2$ and loss rate. Use hyperparameter optimization to achieve the best fit in terms of $R^2$.
            \item Step 2-1-3: Record the feature importance for each input feature from the Lasso regression model.
            \item Step 2-1-4: Add features with feature importance greater than 0.2 to the list of selected features ($F$).
        \end{itemize}
        \item Step 2-2 (feature selection with all lags): Initialize the input data for feature selection as $I_2=\{\}$ and list of final selected features as $F$. Use the loss rate as the dependent variable.
        \begin{itemize}
            \item Step 2-2-1: For each macroeconomic indicator $i$ $(i=1, ..., n)$, select all lagged values from $I_1$ and append it to $I_2$.
            \item Step 2-2-2: Apply Lasso regression using the input data $I_2$ and loss rate. Use hyperparameter optimization to achieve the best fit in terms of $R^2$.
            \item Step 2-2-3: Record the feature importance for each input feature from the Lasso regression model.
            \item Step 2-2-4: Add features with feature importance greater than 0.2 to the list of selected features ($F$).
        \end{itemize}

    \end{itemize}
    \item Step 3 (model training): For each machine learning model $j$ $(j=1, ..., m)$:
    \begin{itemize}
        \item Step 3-1: Use input values selected from the feature selection step ($F$) as independent variables and loss rate as the dependent variable.
        \item Step 3-2: Split the data into training and test set.
        \item Step 3-3: Train model $j$ on the training set and test it on the test set. Record $R^2$ for the training set and MSE for both training and test sets. Use hyperparameter optimization to achieve the best fit.
        \item Step 3-4: If $j=m$, compare the performance of all machine learning models and select the best performing one to use in step 4 to generate the final predictions. 
    \end{itemize}
    \item Step 4 (Forecasting): Use the features selected in step 2 ($F$) as input values for the machine learning model selected in step 3. Let t be the number of macroeconomic indicators in $F$, and let $P$ be a list containing the final predictions for macroeconomic indicators in $F$. We will use the MSIC algorithm proposed in \citet{taghiyeh2020forecasting} for the forecasting model selection for each macroeconomic indicator.
    \begin{itemize}
        \item Step 4-1: For each macroeconomic indicator in $F$ $(k=1, ..., t)$:
        \begin{itemize}
            \item Step 4-1-1: Initialize the input data for the MSIC algorithm as $R=\{\}$.
            \item Step 4-1-2: split the time series corresponding to macroeconomic indicator $k$ into 4-year chunks and append it to $R$.
            \item Step 4-1-3: Train the MSIC algorithm on $R$.
            \item Step 4-1-4: Use the entire values for macroeconomic indicator $k$ as input for the MISC to select the best forecasting model and make final predictions for time series $k$. Append the results of the MSIC algorithm to $P$.
        \end{itemize} 
        \item Step 4-2: Use $P$ as the new input to the selected machine learning model in step 3 to generate the final predictions for the loss rate.
    \end{itemize}
\end{itemize}
      
In the next section, we will apply the proposed loss forecasting model on the loss rate data from the top 100 banks in the US from 1985 to 2019.

\section{Numerical Experiments}

In this section, we will test the proposed loss forecasting framework on the Charge-off rate data from the first quarter of 1985 to the second quarter of 2019. This data is retrieved from the "Board of Governors of the Federal Reserve System (US)" \citep{CO:2020} database and is an aggregated charge-off report for the top 100 US banks ranked by assets. As we mentioned in the introduction section, this study was originally motivated while the author was working for one the leading credit card issuers in the U.S, and a part of this work was originally developed in that company. However, due to the confidentiality issues, all the data related to the company is omitted in this research and the equivalent publicly accessed datasets were being used as the basis for the numerical experiments. 

To select the macroeconomic indicators for this study, initially the "Principles for navigating the big debt crises" by Ray Dalio \citep{RayDalio2018} was reviewed and the macroeconomic indicators which were mentioned in the book that had a significant correlation with debt, charge-off rate and economic cycles were selected. Several additional macroeconomic indicators were also added to the list using the research articles reviewed in the literature review section. This list was provided to the experts in the leading credit card company, including a senior manager and a director from credit risk assessment department. These experts provided their feedback on these indicators and selected 19 indicators that they believed are the ones having the most significant relationship with the charge-off rate and cover all aspects of the economy, while having the smallest overlap to reduce the risk of overfitting. The list of selected macroeconomic indicators is shown in table \ref{table:Indicators}. Please refer to Table A.1 in the appendix for the list of references corresponding to each macroeconomic indicator.

\begin{table}[]
\caption{List of macroeconomic indicators used in this study for building the loss forecasting framework.}
\label{table:Indicators}
\begin{tabular}{|c|ccc}
\hline
\textbf{\begin{tabular}[c]{@{}c@{}}Consumer Segment \\ (Part 1)\end{tabular}} & \multicolumn{1}{c|}{\textbf{\begin{tabular}[c]{@{}c@{}}Consumer Segment \\ (Part 2)\end{tabular}}} & \multicolumn{1}{c|}{\textbf{\begin{tabular}[c]{@{}c@{}}Business \\ Segment\end{tabular}}} & \multicolumn{1}{c|}{\textbf{\begin{tabular}[c]{@{}c@{}}Government \\ Segment\end{tabular}}} \\ \hline
Building Permits& \multicolumn{1}{c|}{\begin{tabular}[c]{@{}c@{}}S\&P 500  \\ Index \end{tabular}} & \multicolumn{1}{c|}{\begin{tabular}[c]{@{}c@{}}Industrial Production \\ Index\end{tabular}} & \multicolumn{1}{c|}{M1} \\ \hline
Housing Starts & \multicolumn{1}{c|}{\begin{tabular}[c]{@{}c@{}}Dow Jones Industrial \\ Average  \end{tabular}} & \multicolumn{1}{c|}{\begin{tabular}[c]{@{}c@{}}ISM Manufacturing\\ New Orders \end{tabular}} & \multicolumn{1}{c|}{M2 } \\ \hline
\begin{tabular}[c]{@{}c@{}}Initial Unemployment \\ Insurance Claims \end{tabular} & \multicolumn{1}{c|}{\begin{tabular}[c]{@{}c@{}}Total Credit \\ Utilization\end{tabular}} & \multicolumn{1}{c|}{\begin{tabular}[c]{@{}c@{}}ISM Purchasing \\ Mangers Index (PMI) \end{tabular}} & \multicolumn{1}{c|}{\begin{tabular}[c]{@{}c@{}}Yield (10 years\\  minus 3 month)\end{tabular}} \\ \hline
Unemployment Rate  & \multicolumn{1}{c|}{\begin{tabular}[c]{@{}c@{}}Revolving Credit \\ Utilization \end{tabular}} & \multicolumn{1}{c|}{\begin{tabular}[c]{@{}c@{}}Weekly Hours\\  Worked by \\ Manufacturing Workers\end{tabular}} & \multicolumn{1}{c|}{\begin{tabular}[c]{@{}c@{}}Yield (10 years\\  minus Federal\\  Fund Rate) \end{tabular}} \\ \hline
\begin{tabular}[c]{@{}c@{}}Consumer Confidence\\  Index (CCI) \end{tabular} & \multicolumn{1}{c|}{\begin{tabular}[c]{@{}c@{}}Non Revolving \\ Credit Utilization \end{tabular}} &  &  \\ \cline{1-2}
\begin{tabular}[c]{@{}c@{}}University of Michigan\\  Sentiment Index \end{tabular} &  &  &  \\ \cline{1-1}
\end{tabular}
\end{table}

\begin{figure}[htbp]
        \center{\includegraphics[width=\textwidth]
        {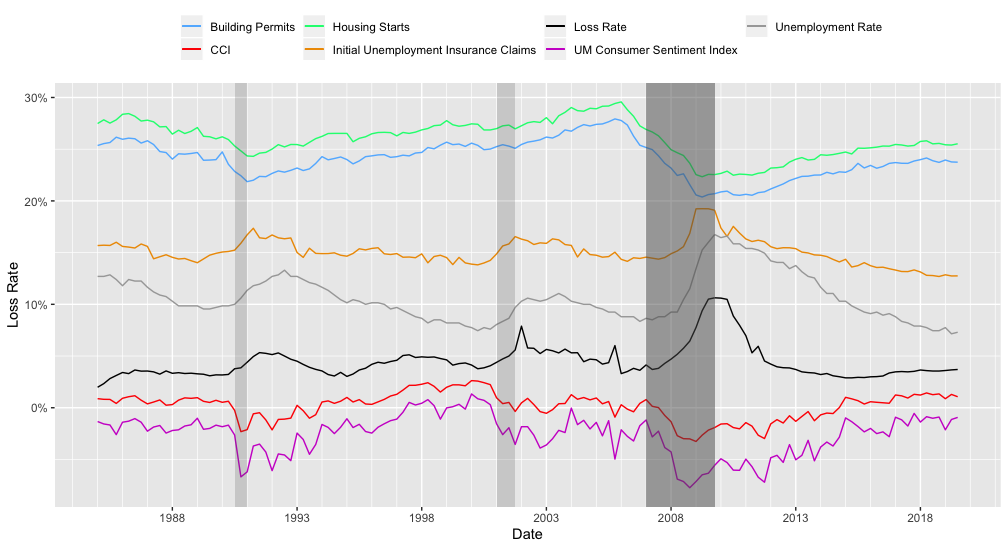}}
        \caption{\label{Households1} Consumer related macroeconomic indicators (part 1)}
      \end{figure}

\begin{figure}[htbp]
        \center{\includegraphics[width=\textwidth]
        {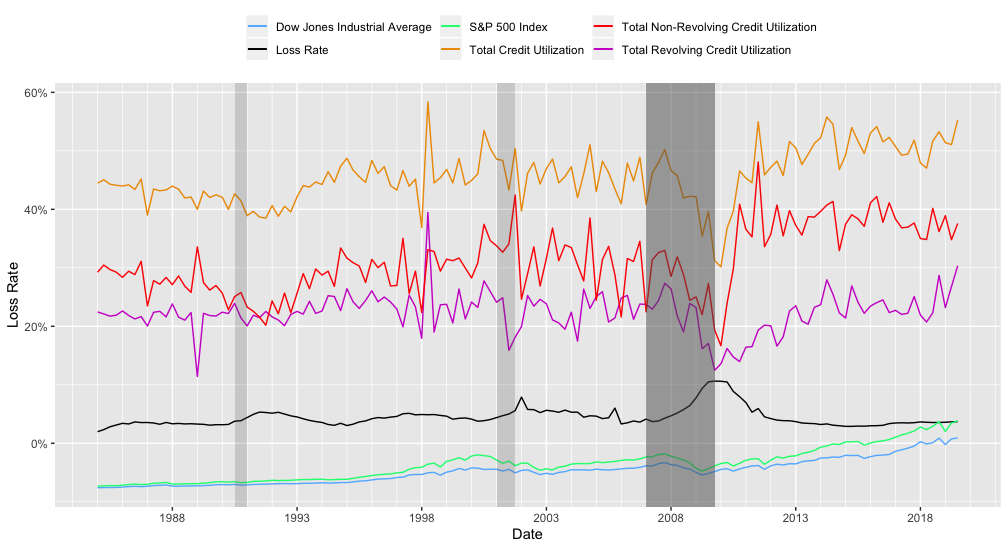}}
        \caption{\label{Households2} Consumer related macroeconomic indicators (part 2)}
      \end{figure}

\begin{figure}[htbp]
        \center{\includegraphics[width=\textwidth]
        {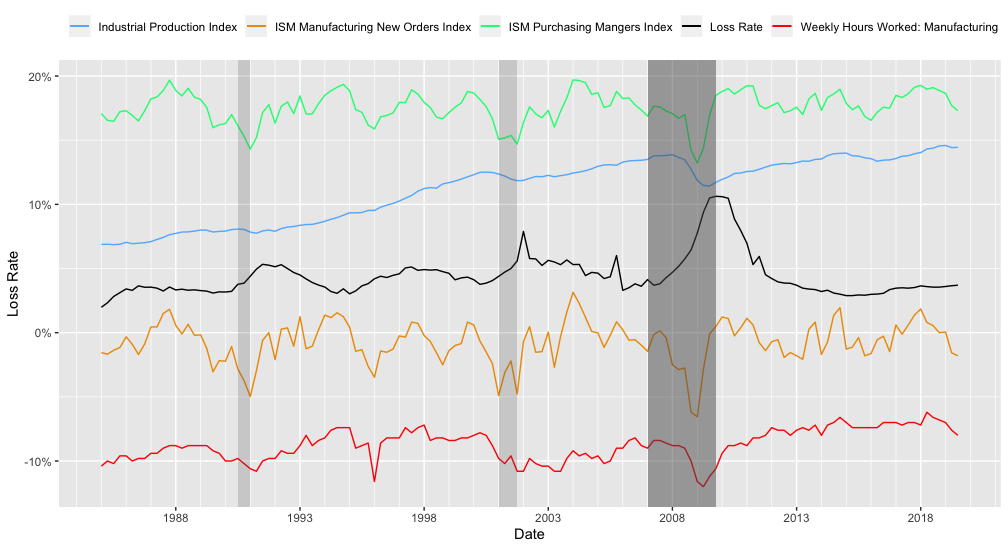}}
        \caption{\label{Business} Manufacturing related macroeconomic indicators}
      \end{figure}

\begin{figure}[htbp]
        \center{\includegraphics[width=\textwidth]
        {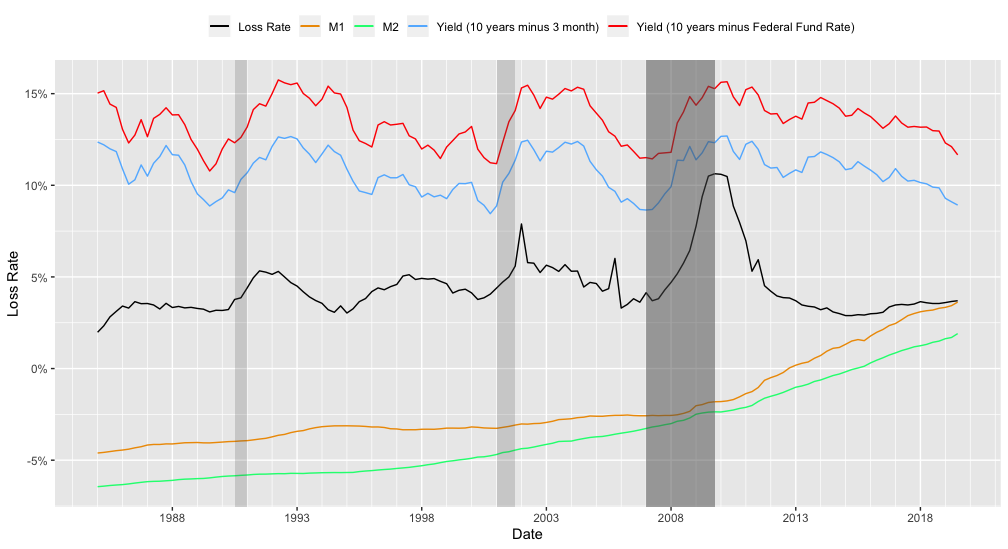}}
        \caption{\label{Government} Government related macroeconomic indicators}
      \end{figure}

Figures \ref{Households1}, \ref{Households2}, \ref{Business}, and \ref{Government} depict the values of the indicators in each segment against charge-off rate. Since each indicator has a different unit, we used the vertical axis to show the charge-off rate, and other indicators were scaled to makes us able to compare their trend against the charge-off rate. The shadowed regions show the US recession periods from 1985 to 2019. As figure \ref{Households1} shows, "unemployment rate" and "initial unemployment insurance claims" have very similar trends to loss rate. That may be the reason that these two indicators are mostly used in the credit card industry to predict the charge-off rate. However, if we look at the values of loss rate, "unemployment rate", and "initial unemployment insurance claims" from the second quarter of 2018 to second quarter of 2019, we see that "unemployment rate" and "initial unemployment insurance claims" are decreasing, but the loss rate has an increasing trend. Hence, there is no way to predict the loss rate in this period by solely using the "unemployment rate" and "initial unemployment insurance claims" as independent variables. Using macroeconomic indicators from all segments of the economy is one of the advantages of our proposed loss forecasting model, which will make it able to capture the uptrend even when the "unemployment rate" and "initial unemployment insurance claims" are decreasing.

Another thing interesting fact is that "M1" and "M2" in table \ref{Government} have significantly different trends before and after the great recession in 2008. As can be seen, the government started printing money in the great recession to add stimulation to the economy and overcome the recession. However, if we look at the trends, they printed money with a significantly higher rate after the great recession, which may be a negative factor for the economy and could play an important role in our loss forecasting framework when we train the model. Moreover, "building permits" and "housing starts" have a very similar trend in figure \ref{Households1}, and to avoid overfitting, only one of them needs to be selected for building a prediction model. The same is true for "CCI" and "UM consumer sentiment index" in figure \ref{Households1}, "Dow Jones industrial average" and "S\&P 500 index" in figure \ref{Households2}, and "yield (10 year minus 3 month" and "yield (10 year minus federal fund rate" inf figure \ref{Government}. These collinearities will be handled by the feature selection step of our proposed loss forecasting model. We will show the step by step implementation of our proposed loss forecasting framework in the following subsections.

\subsection{Data Preparation}
All the macroeconomic indicators are converted to quarterly values, and the lagged values are recorded (1 to 4 quarters). Hence, for each macroeconomic indicator, we have five columns of input data, and in total, we have 95 input columns for 19 macroeconomic indicators in this study. "bestNormalize" package in R \citep{peterson2017bestnormalize} is used for the normalization of each lagged input. The function "bestNormalize" in the aforementioned package performs several normalization transformations, including the Box-Cox transformation, the Yeo-Johnson transformation, the square-root transformation, log transformation, and arcsinh transformation, and uses the Pearson P test statistic for normality to select the optimal one. After performing the optimal transformation selected by "bestNormalize" function, we convert all the values for macroeconomic indicators and the loss rate to year over year changes by dividing them by corresponding values from the previous year. Now we have the input data ready for feature selection step.

\subsection{Feature Selection}
Using the input data obtained from the data preparation step, we start performing the two versions of our feature selection procedures, "feature selection with optimal lags" and "feature selection with all lags".

\subsubsection{Feature Selection with Optimal Lags}
To use the feature selection with optimal lags, we first need to find the optimal lag from the input data generated in step 1. We calculated the correlations of lagged values for each macroeconomic indicator and selected the lag with the highest correlation for each one. The results are shown in table \ref{tab:correlation_analysis}. As we can see, "initial unemployment insurance claims" and "unemployment rate" have the highest correlations with the loss rate, which is in line with what we have already seen in figure \ref{Households1}. Now we perform Lasso regression on these optimal lags to remove collinearity between variables and select the most significant features among the indicators list in table \ref{tab:correlation_analysis}. We used the feature importance list from the results of Lasso regression and selected the indicators with the importance values greater than 0.2. The results for this feature selection procedure is shown in table \ref{tab:Lasso_Optimal}.

\begin{table}[htbp]
  \centering
  \caption{Results of correlation analysis and their statistical significance for economic indicators using different lags}
    \begin{adjustbox}{width=\textwidth}

    \begin{tabular}{|c|c|c|c|c|}
    \hline
    \textbf{Indicators} & \textbf{Lag} & \textbf{Correlation} & \textbf{P-Values} & \textbf{Significance at $\alpha=0.1$} \\
    \hline
    Building Permits & 1     & -0.35 & 0.0212 & Yes \\
    \hline
    CCI   & 2     & -0.49 & 0.0466 & Yes \\
    \hline
    Dow Jones Industrial Average & 3     & -0.09 & 0.7321 & No \\
    \hline
    Housing Starts & 0     & 0.37  & 0.8940 & No \\
    \hline
    Industrial Production Index & 0     & 0.18  & 0.2500 & No \\
    \hline
    Initial Unemployment Insurance Claims & 1     & 0.75  & 0.0038 & Yes \\
    \hline
    M1    & 4     & -0.3  & 0.0659 & Yes \\
    \hline
    M2    & 4     & -0.17 & 0.0675 & Yes \\
    \hline
    ISM Manufacturing New Orders & 4     & -0.46 & 0.0550 & Yes \\
    \hline
    PMI   & 4     & -0.48 & 0.0098 & Yes \\
    \hline
    S\&P 500 Index & 3     & -0.1  & 0.6452 & No \\
    \hline
    University of Michigan Sentiment Index & 2     & -0.48 & 0.0542 & Yes \\
    \hline
    Weekly Hours Worked by Manufacturing Orders & 2     & -0.53 & 0.0360 & Yes \\
    \hline
    Yield (10 years minus 3 months) & 0     & 0.39  & 0.9602 & No \\
    \hline
    Yield (10 years minus Federal Fund Rate) & 0     & 0.39  & 0.2345 & No \\
    \hline
    Unemployment Rate & 0     & 0.52  & 0.0776 & Yes \\
    \hline
    Total Credit Utilization & 0     & -0.49 & 0.7390 & No \\
    \hline
    Revolving Credit Utilization & 0     & -0.48 & 0.3667 & No \\
    \hline
    Non Revolving Credit utilization & 2     & -0.39 & 0.7898 & No \\
    \hline
    \end{tabular}%
  \label{tab:correlation_analysis}%
  \end{adjustbox}
\end{table}%

\begin{table}[htbp]
  \centering
  \caption{Selected indicators using Lasso regression and optimal lags}
    \begin{tabular}{|c|c|c|c|}
\hline
    Indicators & Lag   & Correlation & Relative Importance \\
\hline
    Building Permits & 1     & -0.351927 & 0.84 \\
\hline
    Initial Unemployment Insurance Claims & 1     & 0.74811298 & 0.99 \\
\hline
    M1    & 4     & -0.2975332 & 0.22 \\
\hline
    PMI   & 4     & -0.4789845 & 0.45 \\
\hline
    Weekly Hours Worked by Manufacturing Workers & 0     & -0.5314578 & 0.48 \\
\hline
    Unemployment Rate & 1     & 0.52158091 & 1 \\
    \hline
    \end{tabular}%
  \label{tab:Lasso_Optimal}%
\end{table}%

As we can see in table \ref{tab:Lasso_Optimal}, only six macroeconomic indicators among the initial 19 indicators are selected using feature selection with optimal lags. These macroeconomic indicators are "buliding permits", "initial unemployment insurance claims", "M1", "PMI", "Weekly hours worked by manufacturing workers", and "unemployment rate". If we look at table \ref{table:Indicators}, we can see the interesting result that these indicators cover all the segments mentioned in the table. "Building permits", "Initial unemployment insurance claims" and "unemployment rate" are from the consumer segment. "PMI" and "weekly hours worked by manufacturing workers" are from the business segment, and "M1" covers the government segment of the economy. The fact that our feature selection procedure selected indicators from all segments of the economy suggests that a holistic view of the economy is a requirement to build an effective loss forecasting framework. Additionally, we can see that "M1" is selected as a significant factor, and is in line with what we already suspected as the trend of "M1" is changed significantly after the great recession.

\subsubsection{Feature Selection with All Lags}
As opposed to the feature selection with optimal lags, in this version of feature selection, we do not select the optimal lags manually. We feed all the lagged values of macroeconomic indicators (95 input columns) to the model and let the model itself select the lagged indicators that are the most significant to predict loss. We applied Lasso regression on the input data and selected the indicators according to their relative importance. Lagged indicators with the relative importance greater than 0.2 are selected as final selection for the next step. The results are shown in table \ref{tab:Lasso_All}. As we can see, the selected indicators are almost the same as what we have in feature selection with optimal lags, and all the indicators from feature selection with optimal lags (table \ref{tab:Lasso_Optimal} are selected along with M2. Again, these macroeconomic indicators cover all the segments of the economy (consumer, business, and government segments). The main difference between the selected features in this version is that we allow multiple lags for one indicator to be selected. This way, the final model can also capture the trend of these macroeconomic indicators. It is interesting to see that in table \ref{tab:Lasso_All}, for macroeconomic indicators that multiple lags are selected, these lags have at least two quarters difference. It means that the feature selection procedure tries to capture the most information by using the least number of variables in the cases that the trend had an important role.

\begin{table}[htbp]
  \centering
  \caption{Selected indicators using Lasso regression and all lags}
    \begin{tabular}{|c|c|c|c|}
    \hline
    Indicators & Lag   & Correlation & Relative Importance \\
    \hline
    Building Permits & 1     & -0.351183 & 0.39 \\
    \hline
    Initial Unemployment Insurance Claims & 0     & 0.7169161 & 0.93 \\
    \hline
    Initial Unemployment Insurance Claims & 2     & 0.7262981 & 0.55 \\
    \hline
    Initial Unemployment Insurance Claims & 4     & 0.570129 & 0.71 \\
    \hline
    M1    & 1     & -0.187398 & 0.47 \\
    \hline
    M2    & 4     & -0.165263 & 0.38 \\
    \hline
    PMI   & 0     & -0.431299 & 1 \\
    \hline
    PMI   & 2     & -0.36955 & 0.48 \\
    \hline
    Weekly Hours Worked by Manufacturing Workers & 0     & -0.436368 & 0.47 \\
    \hline
    Unemployment Rate & 0     & 0.5215809 & 0.66 \\
    \hline
    Unemployment Rate & 4     & 0.1248201 & 0.92 \\
    \hline
    \end{tabular}%
  \label{tab:Lasso_All}%
\end{table}%

\subsection{Model Training}
\begin{figure}[htbp]
        \center{\includegraphics[width=\textwidth]
        {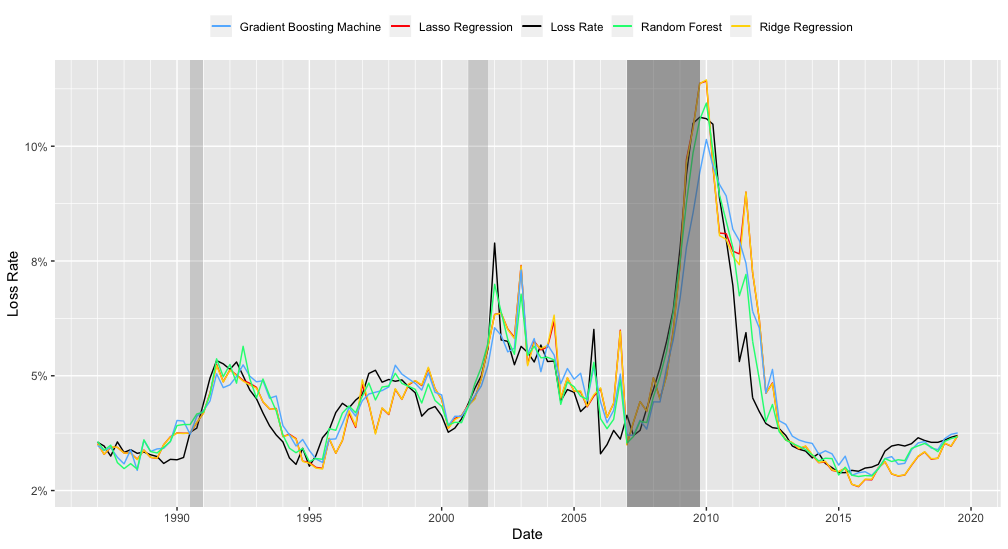}}
        \caption{\label{fig:Optimal_Lags} Final fits for machine learning models using optimal lags as input variables}
      \end{figure}

The features selected by each of our feature selection procedures will be used as input to our machine learning models. The benchmark machine learning models that we use in this study are Lasso regression, Ridge regression, gradient boosting machine, and random forest. We use the data from the first quarter of 2011 to the second quarter of 2019 as the test set and develop two sets of results corresponding to each of our feature selection procedures. We report $R^2$ for the training set and Mean Squared Error (MSE) for both training and test sets. The results using the output of "feature selection with optimal lags" are reported in table \ref{tab:stats_optimal}. The corresponding plots for the fit of each machine learning model are shown in figure \ref{fig:Optimal_Lags}. Comparing the values of $R^2$ in table \ref{tab:stats_optimal}, we see that the gradient boosting machine shows a better performance in terms of $R^2$, which means that 77\% of variations in the loss rate can be explained by the gradient boosting method using optimal lags. The values of MSE in training and test sets are also in line with our conclusion, and the gradient boosting machine shows the best performance in terms of MSE on both training and test sets. Hence, the gradient boosting machine is selected for making the final prediction in the next step when we use Lasso with optimal lags as our model selection procedure.

\begin{table}[htbp]
  \centering
  \caption{Summary statistics for machine learning models when using indicators with optimal lags}
    \begin{tabular}{|c|c|c|c|}
\hline
     & \textbf{$R^2$} & \textbf{MSE (Train)} & \textbf{MSE (Validation)} \\
\hline
    Lasso Regression & 0.72  & 1.59E-02 & 1.90E-02 \\
\hline
    Ridge Regression & 0.72  & 1.14E-02 & 1.55E-02 \\
\hline
    Gradient Boosting Machine & 0.77  & 4.43E-03 & 7.21E-03 \\
\hline
    Random Forest & 0.7   & 1.60E-02 & 1.86E-02 \\
    \hline
    \end{tabular}%
  \label{tab:stats_optimal}%
\end{table}%

\begin{table}[htbp]
  \centering
  \caption{Coefficients and relative importance for machine learning models when using indicators with optimal lags.}
  \begin{adjustbox}{width=\textwidth}
    \begin{tabular}{|c|c|c|c|c|c|}
\cmidrule{3-6}    \multicolumn{1}{c}{} &       & \multicolumn{2}{c|}{\textbf{Coefficients}} & \multicolumn{2}{c|}{\textbf{Relative Importance}} \\
    \hline
    \textbf{Indicator} & \textbf{Lag} & \multicolumn{1}{p{12.665em}|}{\textbf{Lasso Regression}} & \multicolumn{1}{p{12.75em}|}{\textbf{ Ridge Regression}} & \multicolumn{1}{p{11.5em}|}{\textbf{Gradient Boosting \newline{}Machine}} & \textbf{Random Forest} \\
    \hline
    Intercept & -     & 0.0388 & 0.0387 & -     & - \\
    \hline
    Building Permits & 1     & -0.3915 & -0.3922 & 0.1528 & 0.1528 \\
    \hline
    Initial Unemployment Insurance Claims & 1     & 0.3750 & 0.3585 & 1.0000 & 1.0000 \\
    \hline
    M1    & 4     & 0.0112 & 0.0135 & 0.0422 & 0.0422 \\
    \hline
    PMI   & 4     & -0.2070 & -0.2094 & 0.0387 & 0.0387 \\
    \hline
    Weekly Hours Worked by Manufacturing Orders & 2     & -2.2712 & -2.5655 & 0.0508 & 0.0508 \\
    \hline
    Unemployment Rate & 0     & 0.3979 & 0.3915 & 0.2061 & 0.2061 \\
    \hline
    \end{tabular}%
    \end{adjustbox}
  \label{tab:coefficients_Optimal}%
\end{table}%

\begin{figure}[htbp]
        \center{\includegraphics[width=\textwidth]
        {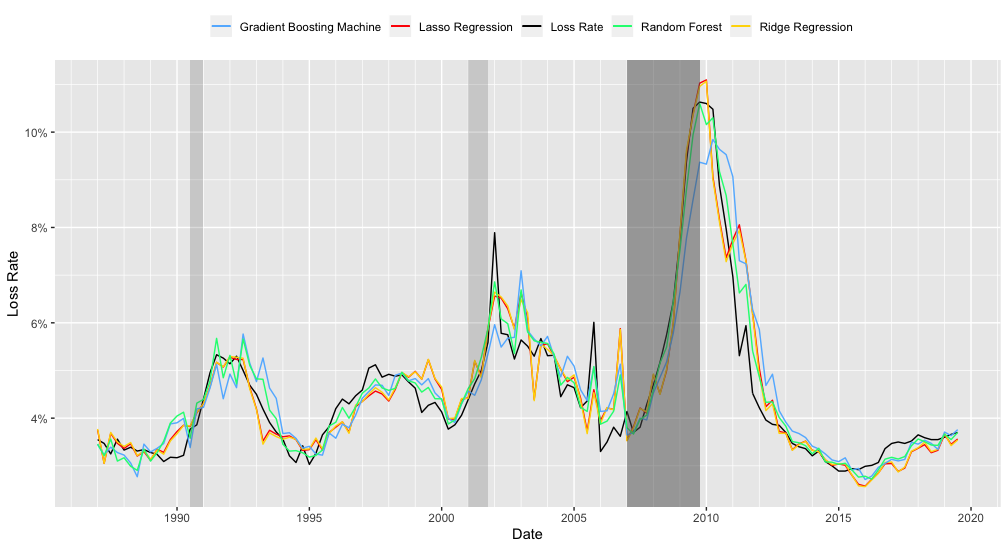}}
        \caption{\label{fig:All_Lags} Final fits for machine learning models using all lags as input variables}
      \end{figure}

Table \ref{tab:stats_all} shows the statistics corresponding to the result of each machine learning method when using feature selection with all lags. The final fit for each method is depicted in figure \ref{fig:All_Lags}. The $R^2$ results in table \ref{fig:All_Lags} suggest that both Lasso and Ridge regression have similar performance. However, looking at the values of MSE in the training and test sets, we see that Ridge regression has a better performance on both train and validation sets. Hence, we select Ridge regression for generating the final forecasts when we use the output of feature selection with all lags as the input of the machine learning model. 

As figures \ref{fig:Optimal_Lags} and \ref{fig:All_Lags} show, the uptrend of the loss rate in the last four quarters can be captured by all the benchmark models using selected features, which is not possible when the unemployment rate is the only decision variable. Additionally, we can see that all the models are able to capture the trends of loss rate with an acceptable accuracy, which shows that our loss forecasting method can provide acceptable results using any of the benchmark machine learning models. We use the selected machine learning model in this step to generate final forecasts in the next step of our loss forecasting algorithm, which is explained in the next subsection.      

\begin{table}[htbp]
  \centering
  \caption{Summary statistics for machine learning models when using indicators with all lags}
    \begin{tabular}{|c|c|c|c|}
\hline
    \textbf{Model 2: All Lags} & \textbf{$R^2$} & \textbf{MSE (Train)} & \textbf{MSE (Validation)} \\
\hline
    Lasso Regression & 0.81  & 9.62E-03 & 1.20E-02 \\
\hline
    Ridge Regression & 0.81  & 3.82E-03 & 7.85E-03 \\
\hline
    Gradient Boosting Machine & 0.77  & 1.01E-02 & 1.20E-02 \\
\hline
    Random Forest & 0.72  & 1.05E-02 & 1.67E-02 \\
    \hline
    \end{tabular}%
  \label{tab:stats_all}%
\end{table}%

\begin{table}[htbp]
  \centering
  \caption{Coefficients and relative importance for machine learning models when using indicators with all lags.}
  \begin{adjustbox}{width=\textwidth}
    \begin{tabular}{|c|c|c|c|c|c|}
\cmidrule{3-6}    \multicolumn{1}{r}{} &       & \multicolumn{2}{c|}{\textbf{Coefficients}} & \multicolumn{2}{c|}{\textbf{Relative Importance}} \\
    \hline
    \textbf{Indicator} & \textbf{Lag} & \textbf{Lasso Regression} & \textbf{Ridge Regression} & \multicolumn{1}{p{11.5em}|}{\textbf{Gradient Boosting \newline{}Machine}} & \textbf{Random Forest} \\
    \hline
    Intercept & -     & 0.0265 & 0.0250 & -     & - \\
    \hline
    Building Permits & 1     & -0.2848 & -0.2882 & 0.1552 & 0.5193 \\
    \hline
    Initial Unemployment Insurance Claims & 0     & 0.5631 & 0.5355 & 0.7668 & 1.0000 \\
    \hline
    Initial Unemployment Insurance Claims & 2     & 0.3184 & 0.3531 & 0.3061 & 0.5519 \\
    \hline
    Initial Unemployment Insurance Claims & 4     & 0.4846 & 0.4877 & 0.1558 & 0.1674 \\
    \hline
    M1    & 1     & 0.0267 & 0.0295 & 0.0436 & 0.0787 \\
    \hline
    M2    & 4     & 0.0098 & 0.0157 & 0.0291 & 0.0655 \\
    \hline
    PMI   & 0     & 0.5100 & 0.5518 & 0.0325 & 0.0979 \\
    \hline
    PMI   & 2     & 0.0549 & 0.1134 & 0.0544 & 0.1976 \\
    \hline
    Weekly Hours Worked by Manufacturing Orders & 0     & -2.9343 & -3.1827 & 0.0238 & 0.1245 \\
    \hline
    Unemployment Rate & 0     & 0.3452 & 0.4272 & 1.0000 & 0.6851 \\
    \hline
    Unemployment Rate & 4     & 0.3910 & 0.4802 & 0.0166 & 0.0764 \\
    \hline
    \end{tabular}%
    \end{adjustbox}
  \label{tab:coefficients_all}%
\end{table}%

\subsection{Forecasting}
The last step to build the loss forecasting framework is to predict each macroeconomic indicator and use the trained model in step 3 to predict the future values of the loss rate. We use the second quarter of 2018 to the second quarter of 2019 (1 year) as the prediction period. We predict each macroeconomic indicator for this period, and using the trained model in step 3, we will forecast the loss rate. We will use actual values for this period to evaluate the forecasted values. 

To forecast the values of each macroeconomic indicator, we need to select the most appropriate time series forecasting model. Since the performance of forecasting models highly depends on the underlying characteristics of the time series, the selection of the best is not a simple task. As the forecasting model selection approach in \citet{taghiyeh2020forecasting} (MSIC algorithm) has shown promising performance, we will use this procedure to select our forecasting model for each macroeconomic indicator. Similar to \citet{taghiyeh2020forecasting}, we select seven of the most well-known time series forecasting models as our benchmark, namely nai\"ve forecasting, moving average, ARIMA, simple exponential smoothing, Holt's linear trend, Holt-Winters, and theta. For each macroeconomic indicator, the MISC algorithm will select the optimal forecasting model, and we will use the selected optimal model to forecast future values for each macroeconomic indicator.

Since MSIC needs multiple time series as input to train its classifiers, we need to convert the time series associated with each macroeconomic indicator into several series. To achieve this goal, we use non-overlapping four year horizons to split the data for each macroeconomic indicator. We use this input data to train the MSIC classifiers. To make final predictions, we use the entire data for the corresponding macroeconomic indicator as input to the trained classifiers of the MSIC algorithm.

\begin{figure}[htbp]
        \center{\includegraphics[width=\textwidth]
        {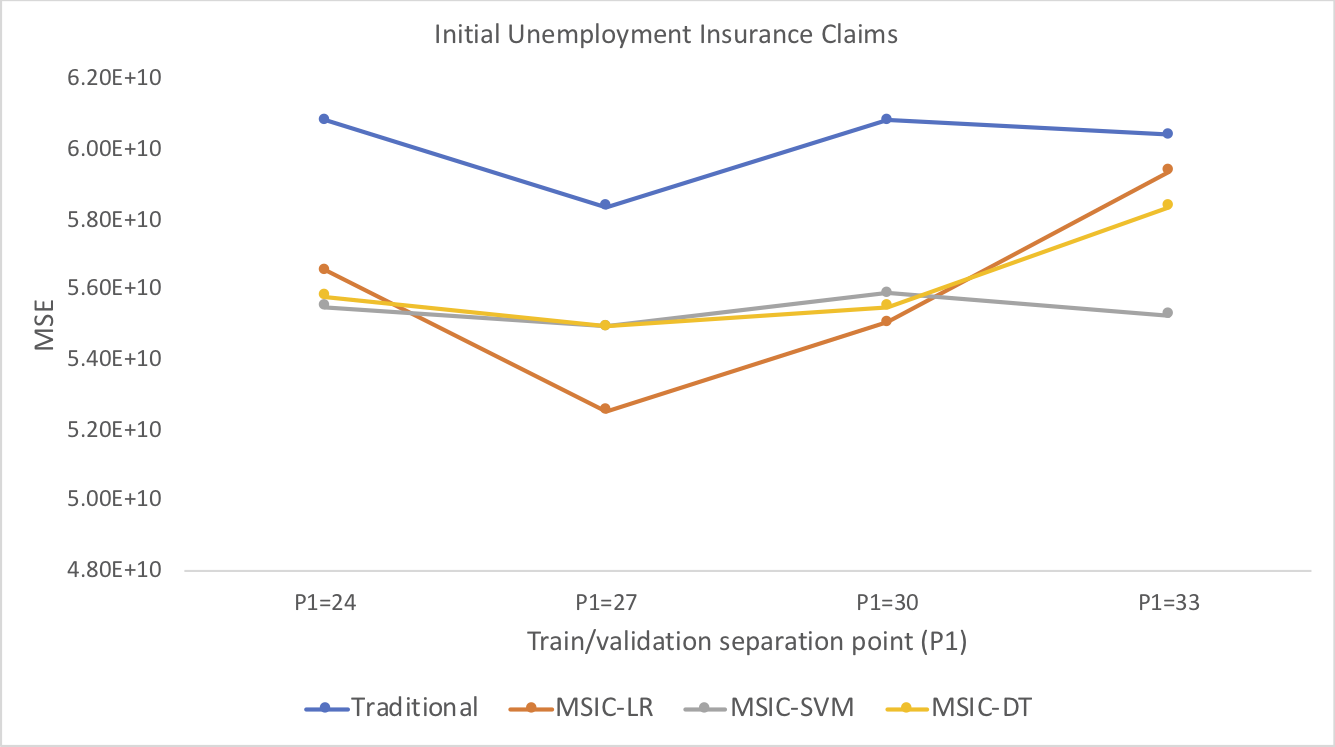}}
        \caption{\label{IUIC} Performance comparison using "Initial Unemployment Insurance Claims" Data}
      \end{figure}

\begin{figure}[htbp]
        \center{\includegraphics[width=\textwidth]
        {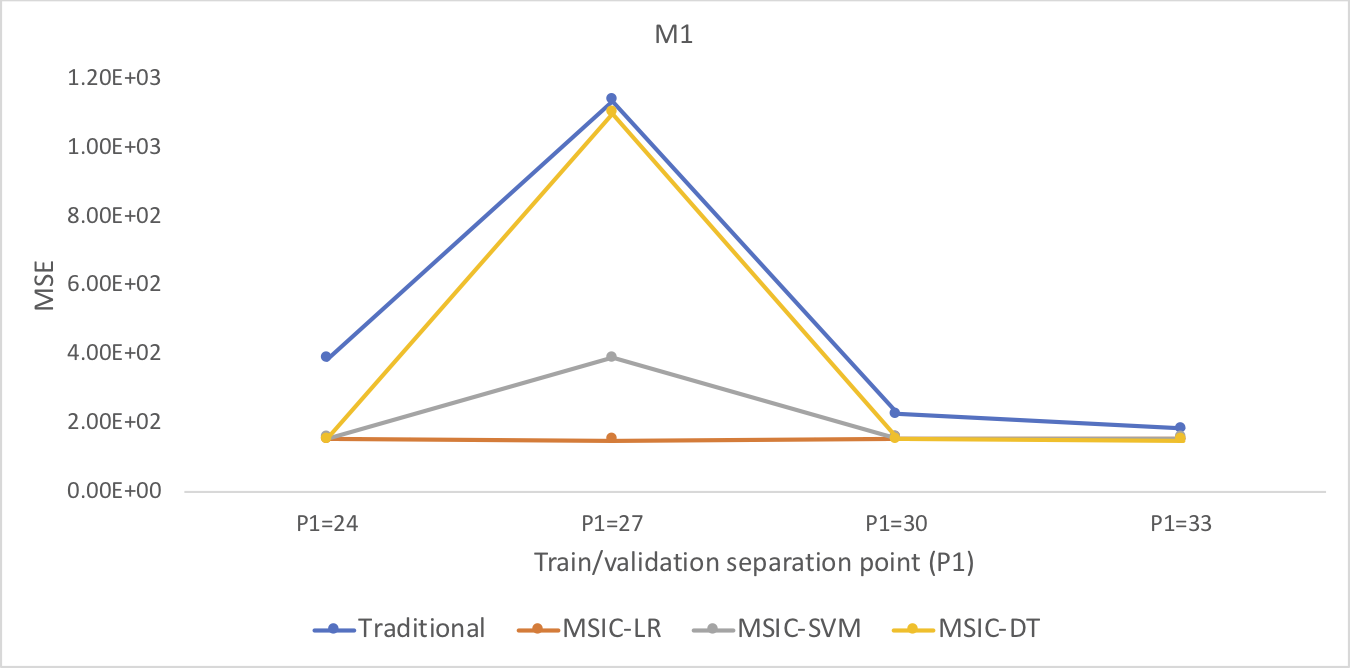}}
        \caption{\label{M1} Performance comparison using "M1" Data}
      \end{figure}

\begin{figure}[htbp]
        \center{\includegraphics[width=\textwidth]
        {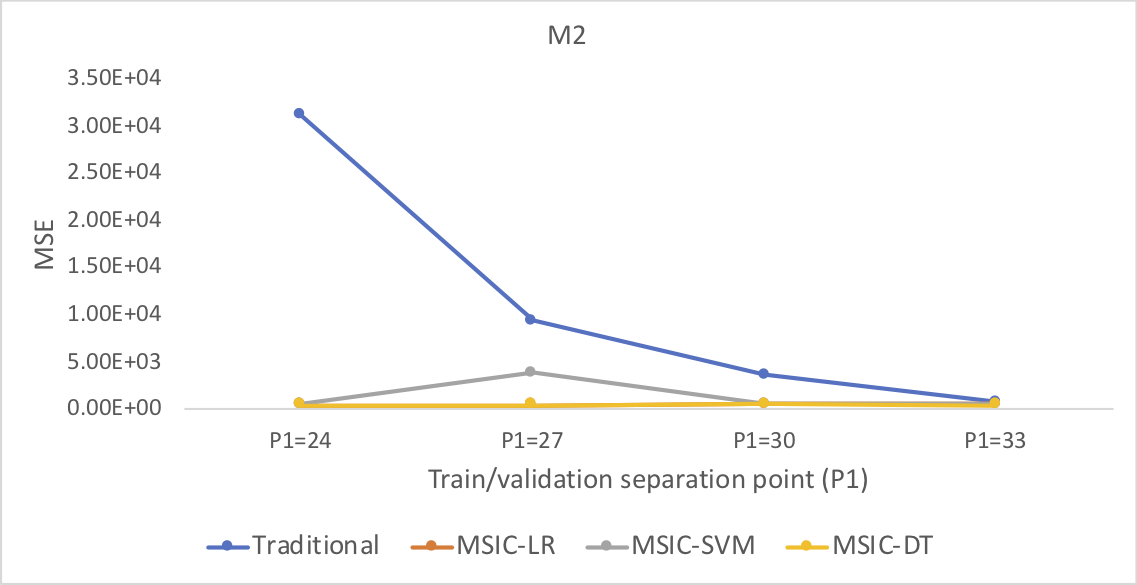}}
        \caption{\label{M2} Performance comparison using "M2" Data}
      \end{figure}

\begin{figure}[htbp]
        \center{\includegraphics[width=\textwidth]
        {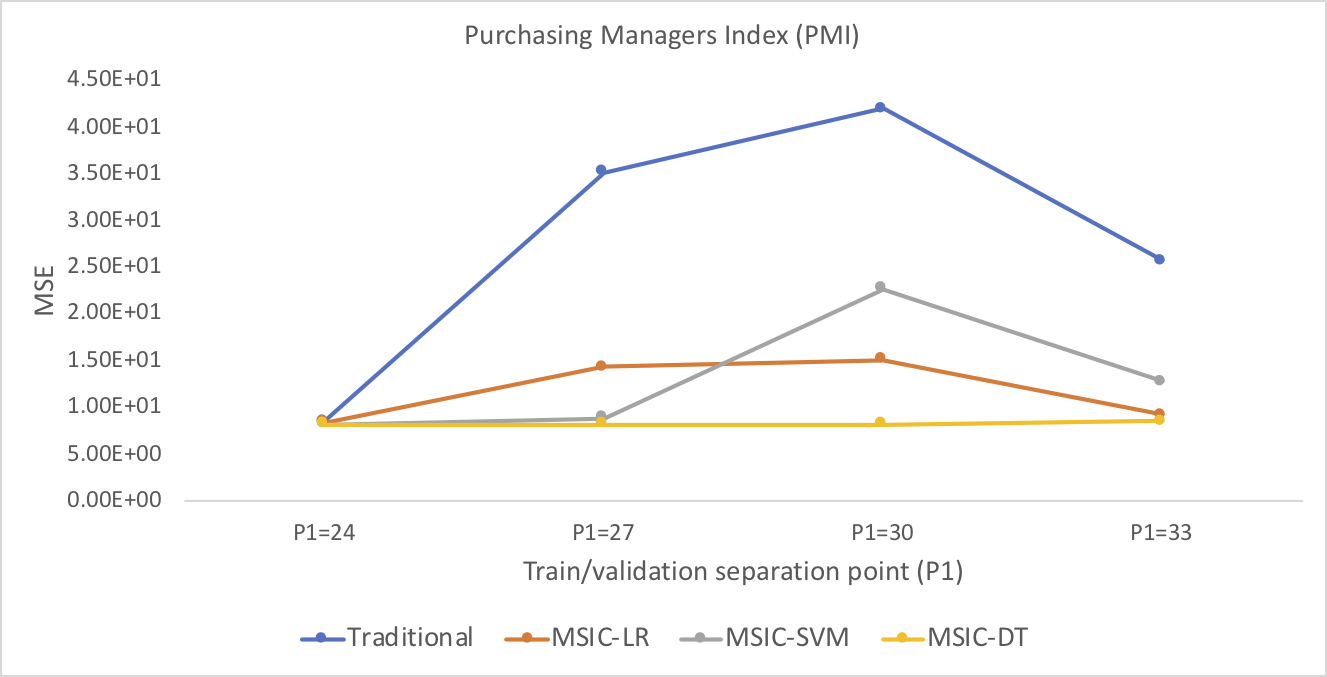}}
        \caption{\label{PMI} Performance comparison using "Purchasing Managers Index (PMI)" Data}
      \end{figure}

\begin{figure}[htbp]
        \center{\includegraphics[width=\textwidth]
        {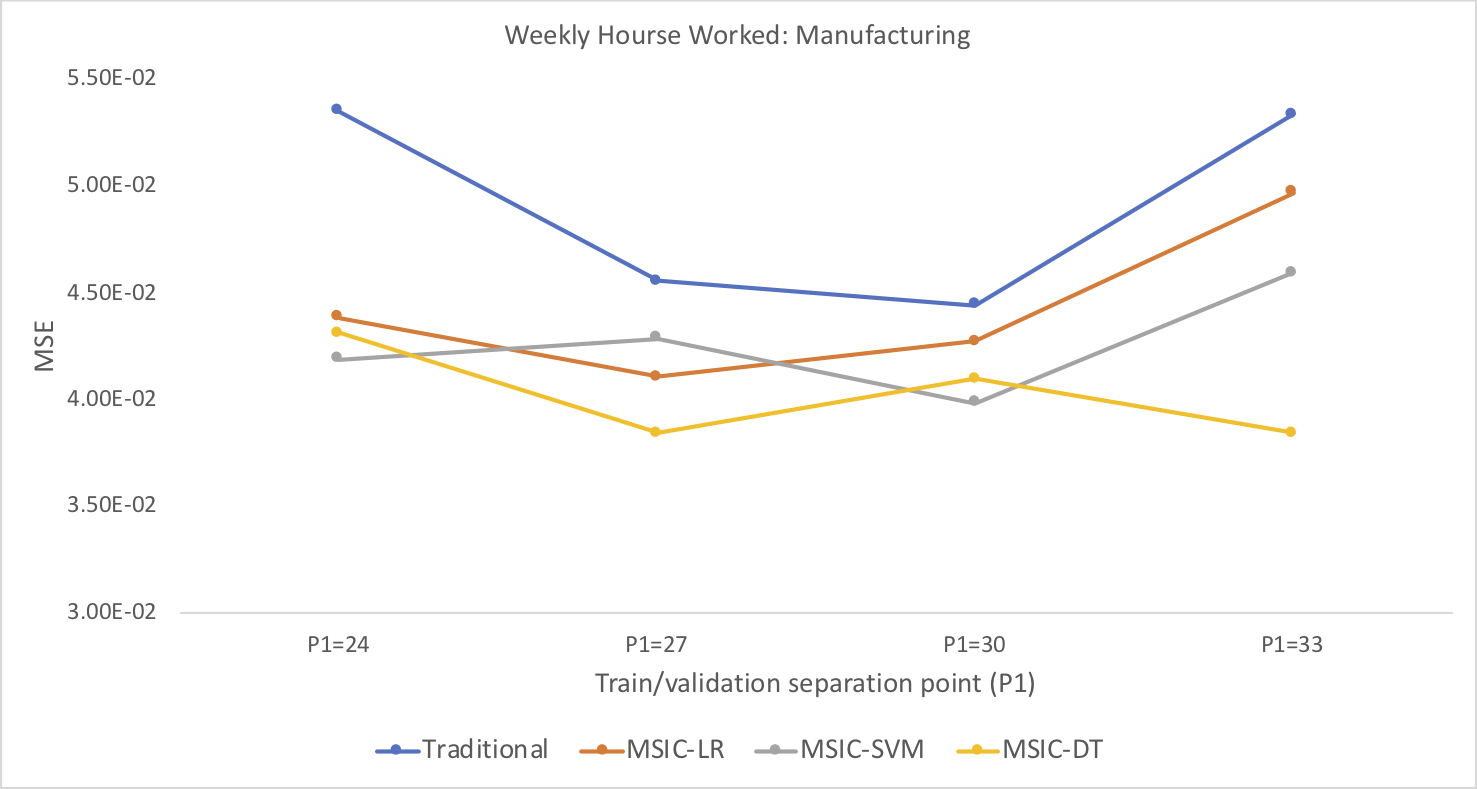}}
        \caption{\label{Weekly} Performance comparison using "Weekly Hours Worked: Manufacturing" Data}
      \end{figure}

\begin{figure}[htbp]
        \center{\includegraphics[width=\textwidth]
        {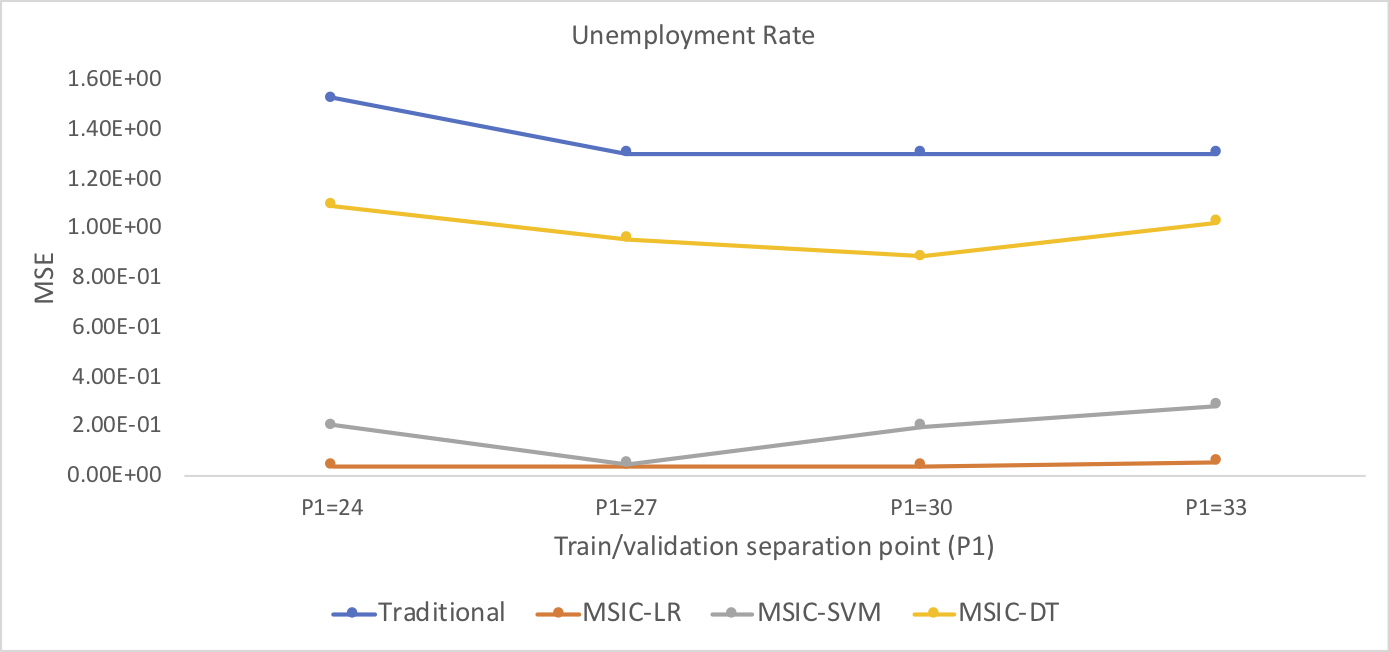}}
        \caption{\label{UR} Performance comparison using "Unemployment Rate" Data}
      \end{figure}

\begin{figure}[htbp]
        \center{\includegraphics[width=\textwidth]
        {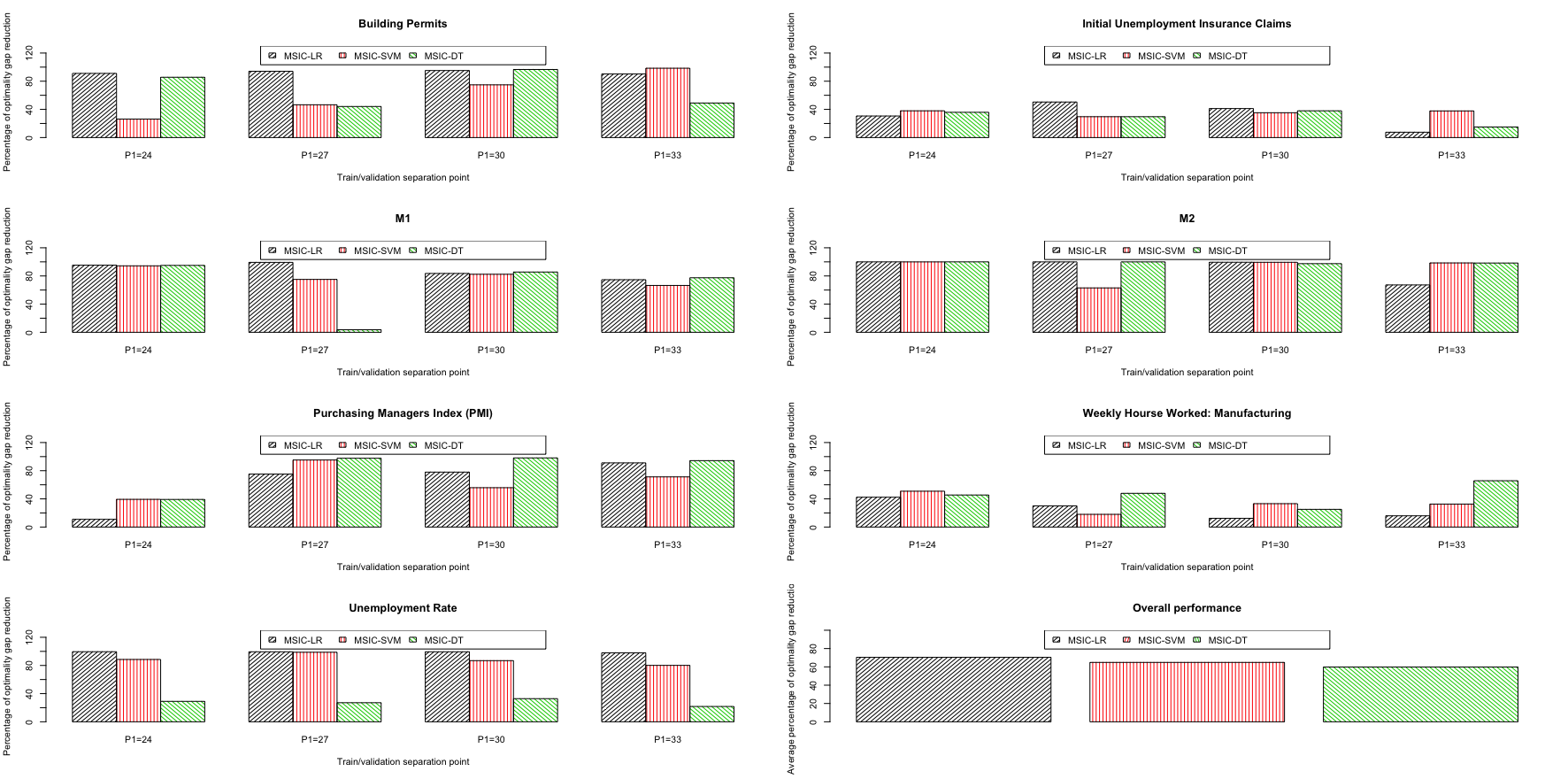}}
        \caption{\label{overal} Optimality gap improvement for all macro economic indicators using three versions of MSIC. Average improvements for all three versions over all categories are shown in last figure. }
      \end{figure}

To evaluate the performance of the MSIC algorithm for each macroeconomic indicator, we compare the results of the MSIC algorithm to the traditional train/validation forecasting model selection method. Three variants of the MSIC algorithm, namely MSIC with logistic regression as the classifier (MSIC-LR), MSIC with support vector machine as a classifier (MSIC-SVM), and MSIC with decision tree as a classifier (MSIC-DT) are used for this comparison, and we report MSE and optimality gap reduction as the comparison measures. To be consistent with the results reported in \citet{taghiyeh2020forecasting}, we use different values for separations points between train and validation sets (P1). Since in the feature selection step (step 2) only 7 of the macroeconomic indicators are selected (building permits, initial unemployment insurance claims, M1, M2, purchasing managers index, weekly hours worked by manufacturing workers and unemployment rate), we only use the MSIC algorithm to predict future values for these indicators. The comparison results for the selected macroeconomic indicators are shown in tables \ref{tab:BP}--\ref{tab:UR}. The MSE results are also depicted in figures \ref{BP}--\ref{UR}. The optimality gap improvements are summarized in figure \ref{overal}. 

The results suggest the same trend as numerical results in \citet{taghiyeh2020forecasting}, as the MSIC algorithm shows a constant improvement in the optimality gap in all instances. Additionally, there is not a single winner among classifiers for the MSIC algorithm, and it is case dependent. As the overall performance in figure \ref{overal} shows, we can get an overall minimum of 60\% improvement in optimality gap improvement using the MSIC algorithm over the traditional train/validation model selection procedure.

\begin{landscape}
\begin{table}[]
\caption{Comparing the performance of MSIC to traditional train/validation model selection procedure using "Building Permits" data}
\label{tab:BP}
\begin{adjustbox}{width=1.3\textwidth}
\begin{tabular}{|c|c|c|c|c|c|c|c|c|c|}
\hline
\textbf{Building Permits}           & \textbf{Optimal} & \textbf{Traditional} & \multicolumn{2}{c|}{\textbf{MSIC-LR}}         & \multicolumn{2}{c|}{\textbf{MSIC-SVM}}        & \multicolumn{2}{c|}{\textbf{MSIC-DT}}         & \multirow{2}{*}{\textbf{Average Optimal Gap Reduction}} \\ \cline{1-9}
\textbf{Train/validation separation point} & \textbf{MSE}     & \textbf{MSE}         & \textbf{MSE} & \textbf{Optimal Gap Reduction} & \textbf{MSE} & \textbf{Optimal Gap Reduction} & \textbf{MSE} & \textbf{Optimal Gap Reduction} &                                                         \\ \hline
P1=24                                      & 5.86E+03         & 2.17E+04             & 7.27+03     & 91.07\%                        & 1.75E+04     & 26.20\%                        & 8.13E+03     & 85.61\%                        & 67.62\%                                                 \\ \hline
P1=27                                      & 5.86E+03    & 2.97E+04    & 7.27E+03    &94.09\%    &1.86E+04    &46.51\%    &1.92E+04    &44.08\%    &61.56\%                   \\ \hline
P1=30                                      & 5.86E+03    &4.54E+04    &7.77E+03    &95.15\%    &1.58E+04    &74.83\%    &7.18E+03    &96.65\%    &88.88\% \\ \hline
P1=33                                      & 5.86E+03    &1.13E+05    &1.63E+04    &90.26\%    &7.57E+03    &98.41\%    &6.07E+04    &49.02\%    &79.23\%    \\ \hline
\end{tabular}
\end{adjustbox}
\vspace{2cm}
\end{table}

\begin{table}[]
\caption{Comparing the performance of MSIC to traditional train/validation model selection procedure using "Initial Unemployment Insurance Claims" data}
\label{tab:IUIC}
\begin{adjustbox}{width=1.3\textwidth}
\begin{tabular}{|c|c|c|c|c|c|c|c|c|c|}
\hline
\textbf{Initial Unemployment Insurance Claims}               & \textbf{Optimal} & \textbf{Traditional} & \multicolumn{2}{c|}{\textbf{MSIC-LR}}         & \multicolumn{2}{c|}{\textbf{MSIC-SVM}}        & \multicolumn{2}{c|}{\textbf{MSIC-DT}}         & \multirow{2}{*}{\textbf{Average Optimal Gap Reduction}} \\ \cline{1-9}
\textbf{Train/validation separation point} & \textbf{MSE}     & \textbf{MSE}         & \textbf{MSE} & \textbf{Optimal Gap Reduction} & \textbf{MSE} & \textbf{Optimal Gap Reduction} & \textbf{MSE} & \textbf{Optimal Gap Reduction} &                                                         \\ \hline
P1=24                                      & 4.68E+10    &6.08E+10    &5.65E+10    &30.57\%    &5.55E+10    &37.95\%    &5.58E+10    &35.84\%    &34.79\%  \\ \hline
P1=27                                      & 4.68E+10    &5.84E+10    &5.25E+10    &50.43\%    &5.50E+10    &29.59\%    &5.50E+10    &29.59\%    &36.54\% \\ \hline
P1=30                                      & 4.68E+10    &6.08E+10    &5.51E+10    &41.17\%    &5.59E+10    &35.20\%    &5.55E+10    &37.95\%    &38.11\% \\ \hline
P1=33                                      & 4.68E+10    &6.04E+10    &5.94E+10    &7.54\%    &5.53E+10    &37.78\%    &5.84E+10    &14.96\%    &20.09\%    \\ \hline
\end{tabular}
\end{adjustbox}
\vspace{2cm}
\end{table}

\begin{table}[]
\caption{Comparing the performance of MSIC to traditional train/validation model selection procedure using "M1" data}
\label{tab:M1}
\begin{adjustbox}{width=1.3\textwidth}
\begin{tabular}{|c|c|c|c|c|c|c|c|c|c|}
\hline
\textbf{M1}              & \textbf{Optimal} & \textbf{Traditional} & \multicolumn{2}{c|}{\textbf{MSIC-LR}}         & \multicolumn{2}{c|}{\textbf{MSIC-SVM}}        & \multicolumn{2}{c|}{\textbf{MSIC-DT}}         & \multirow{2}{*}{\textbf{Average Optimal Gap Reduction}} \\ \cline{1-9}
\textbf{Train/validation separation point} & \textbf{MSE}     & \textbf{MSE}         & \textbf{MSE} & \textbf{Optimal Gap Reduction} & \textbf{MSE} & \textbf{Optimal Gap Reduction} & \textbf{MSE} & \textbf{Optimal Gap Reduction} &                                                         \\ \hline
P1=24                                      & 1.40E+02    &3.84E+02    &1.51E+02    &95.33\%    &1.54E+02    &94.35\%    &1.52E+02    &94.99\%    &94.89\% \\ \hline
P1=27                                      & 1.40E+02    &1.14E+03    &1.47E+02    &99.28\%    &3.88E+02    &75.08\%    &1.10E+03    &3.60\%    &59.32\%  \\ \hline
P1=30                                      & 1.40E+02    &2.25E+02    &1.54E+02    &83.55\%    &1.55E+02    &82.39\%    &1.52E+02    &85.56\%    &83.83\% \\ \hline
P1=33                                      & 1.40E+02    &1.83E+02    &1.51E+02    &74.65\%    &1.54E+02    &66.52\%    &1.50E+02    &77.43\%    &72.87\%  \\ \hline
\end{tabular}
\end{adjustbox}
\vspace{2cm}
\end{table}

\begin{table}[]
\caption{Comparing the performance of MSIC to traditional train/validation model selection procedure using "M2" data}
\label{tab:M2}
\begin{adjustbox}{width=1.3\textwidth}
\begin{tabular}{|c|c|c|c|c|c|c|c|c|c|}
\hline
\textbf{M2}                 & \textbf{Optimal} & \textbf{Traditional} & \multicolumn{2}{c|}{\textbf{MSIC-LR}}         & \multicolumn{2}{c|}{\textbf{MSIC-SVM}}        & \multicolumn{2}{c|}{\textbf{MSIC-DT}}         & \multirow{2}{*}{\textbf{Average Optimal Gap Reduction}} \\ \cline{1-9}
\textbf{Train/validation separation point} & \textbf{MSE}     & \textbf{MSE}         & \textbf{MSE} & \textbf{Optimal Gap Reduction} & \textbf{MSE} & \textbf{Optimal Gap Reduction} & \textbf{MSE} & \textbf{Optimal Gap Reduction} &                                                         \\ \hline
P1=24                                      & 4.12E+02    &3.12E+04    &4.16E+02    &99.99\%    &4.31E+02    &99.94\%    &4.16E+02    &99.99\%    &99.97\% \\ \hline
P1=27                                      & 4.12E+02    &9.33E+03    &4.16E+02    &99.95\%    &3.71E+03    &63.01\%    &4.16E+02    &99.95\%    &87.64\%  \\ \hline
P1=30                                      & 4.12E+02    &3.52E+03    &4.29E+02    &99.47\%    &4.31E+02    &99.40\%    &4.91E+02    &97.46\%    &98.78\%  \\ \hline
P1=33                                      & 4.12E+02    &6.54E+02    &4.91E+02    &67.23\%    &4.16E+02    &98.45\%    &4.16E+02    &98.32\%    &88.00\%  \\ \hline
\end{tabular}
\end{adjustbox}
\vspace{2cm}
\end{table}

\begin{table}[]
\caption{Comparing the performance of MSIC to traditional train/validation model selection procedure using "Purchasing Managers Index (PMI)" data}
\label{tab:PMI}
\begin{adjustbox}{width=1.3\textwidth}
\begin{tabular}{|c|c|c|c|c|c|c|c|c|c|}
\hline
\textbf{Purchasing Managers Index (PMI)}                 & \textbf{Optimal} & \textbf{Traditional} & \multicolumn{2}{c|}{\textbf{MSIC-LR}}         & \multicolumn{2}{c|}{\textbf{MSIC-SVM}}        & \multicolumn{2}{c|}{\textbf{MSIC-DT}}         & \multirow{2}{*}{\textbf{Average Optimal Gap Reduction}} \\ \cline{1-9}
\textbf{Train/validation separation point} & \textbf{MSE}     & \textbf{MSE}         & \textbf{MSE} & \textbf{Optimal Gap Reduction} & \textbf{MSE} & \textbf{Optimal Gap Reduction} & \textbf{MSE} & \textbf{Optimal Gap Reduction} &                                                         \\ \hline
P1=24                                      & 7.48E+00    &8.52E+00    &8.40E+00    &10.94\%    &8.11E+00    &39.31\%    &8.11E+00    &39.18\%    &29.81\%  \\ \hline
P1=27                                      & 17.48E+00    &3.51E+01    &1.43E+01    &75.22\%    &8.80E+00    &95.19\%    &8.11E+00    &97.70\%    &89.37\%  \\ \hline
P1=30                                      & 7.48E+00    &4.19E+01    &1.51E+01    &77.99\%    &2.27E+01    &55.86\%    &8.11E+00    &98.16\%    &77.34\%  \\ \hline
P1=33                                      & 7.48E+00    &2.57E+01    &9.10E+00    &91.10\%    &1.27E+01    &71.22\%    &8.52E+00    &94.28\%    &85.53\%  \\ \hline
\end{tabular}
\end{adjustbox}
\vspace{2cm}
\end{table}

\begin{table}[]
\caption{Comparing the performance of MSIC to traditional train/validation model selection procedure using "Weekly Hours Worked: Manufacturing" data}
\label{tab:Weekly}
\begin{adjustbox}{width=1.3\textwidth}
\begin{tabular}{|c|c|c|c|c|c|c|c|c|c|}
\hline
\textbf{Weekly Hours Worked: Manufacturing}                 & \textbf{Optimal} & \textbf{Traditional} & \multicolumn{2}{c|}{\textbf{MSIC-LR}}         & \multicolumn{2}{c|}{\textbf{MSIC-SVM}}        & \multicolumn{2}{c|}{\textbf{MSIC-DT}}         & \multirow{2}{*}{\textbf{Average Optimal Gap Reduction}} \\ \cline{1-9}
\textbf{Train/validation separation point} & \textbf{MSE}     & \textbf{MSE}         & \textbf{MSE} & \textbf{Optimal Gap Reduction} & \textbf{MSE} & \textbf{Optimal Gap Reduction} & \textbf{MSE} & \textbf{Optimal Gap Reduction} &                                                         \\ \hline
P1=24                                      & 3.06E-02    &5.35E-02    &4.38E-02    &42.35\%    &4.19E-02    &50.87\%    &4.31E-02    &45.48\%    &46.23\% \\ \hline
P1=27                                      & 3.06E-02    &4.55E-02    &4.11E-02    &30.03\%    &4.28E-02    &18.01\%    &3.84E-02    &47.86\%    &31.97\%  \\ \hline
P1=30                                      & 3.06E-02    &4.44E-02    &4.27E-02    &12.37\%    &3.98E-02    &33.18\%    &4.09E-02    &25.13\%    &23.56\%  \\ \hline
P1=33                                      & 3.06E-02    &5.33E-02    &4.97E-02    &15.95\%    &4.59E-02    &32.60\%    &3.84E-02    &65.74\%    &38.10\%  \\ \hline
\end{tabular}
\end{adjustbox}
\vspace{2cm}
\end{table}

\begin{table}[]
\caption{Comparing the performance of MSIC to traditional train/validation model selection procedure using "Unemployment Rate" data}
\label{tab:UR}
\begin{adjustbox}{width=1.3\textwidth}
\begin{tabular}{|c|c|c|c|c|c|c|c|c|c|}
\hline
\textbf{Unemployment Rate}                 & \textbf{Optimal} & \textbf{Traditional} & \multicolumn{2}{c|}{\textbf{MSIC-LR}}         & \multicolumn{2}{c|}{\textbf{MSIC-SVM}}        & \multicolumn{2}{c|}{\textbf{MSIC-DT}}         & \multirow{2}{*}{\textbf{Average Optimal Gap Reduction}} \\ \cline{1-9}
\textbf{Train/validation separation point} & \textbf{MSE}     & \textbf{MSE}         & \textbf{MSE} & \textbf{Optimal Gap Reduction} & \textbf{MSE} & \textbf{Optimal Gap Reduction} & \textbf{MSE} & \textbf{Optimal Gap Reduction} &                                                         \\ \hline
P1=24                                      & 2.93E-02    &1.52E+00    &3.68E-02    &99.50\%    &2.01E-01    &88.50\%    &1.09E+00    &29.01\%    &72.34\% \\ \hline
P1=27                                      & 2.93E-02    &1.30E+00    &3.68E-02    &99.41\%    &4.53E-02    &98.74\%    &9.58E-01    &27.07\%    &75.07\% \\ \hline
P1=30                                      & 2.93E-02    &1.30E+00    &3.66E-02    &99.43\%    &1.97E-01    &86.83\%    &8.84E-01    &32.84\%    &73.03\%  \\ \hline
P1=33                                      & 12.93E-02    &1.30E+00    &5.54E-02    &97.95\%    &2.83E-01    &80.11\%    &1.03E+00    &21.73\%    &66.60\% \\ \hline
\end{tabular}
\end{adjustbox}
\vspace{2cm}
\end{table}

\begin{figure}[htbp]
        \center{\includegraphics[width=\textwidth]
        {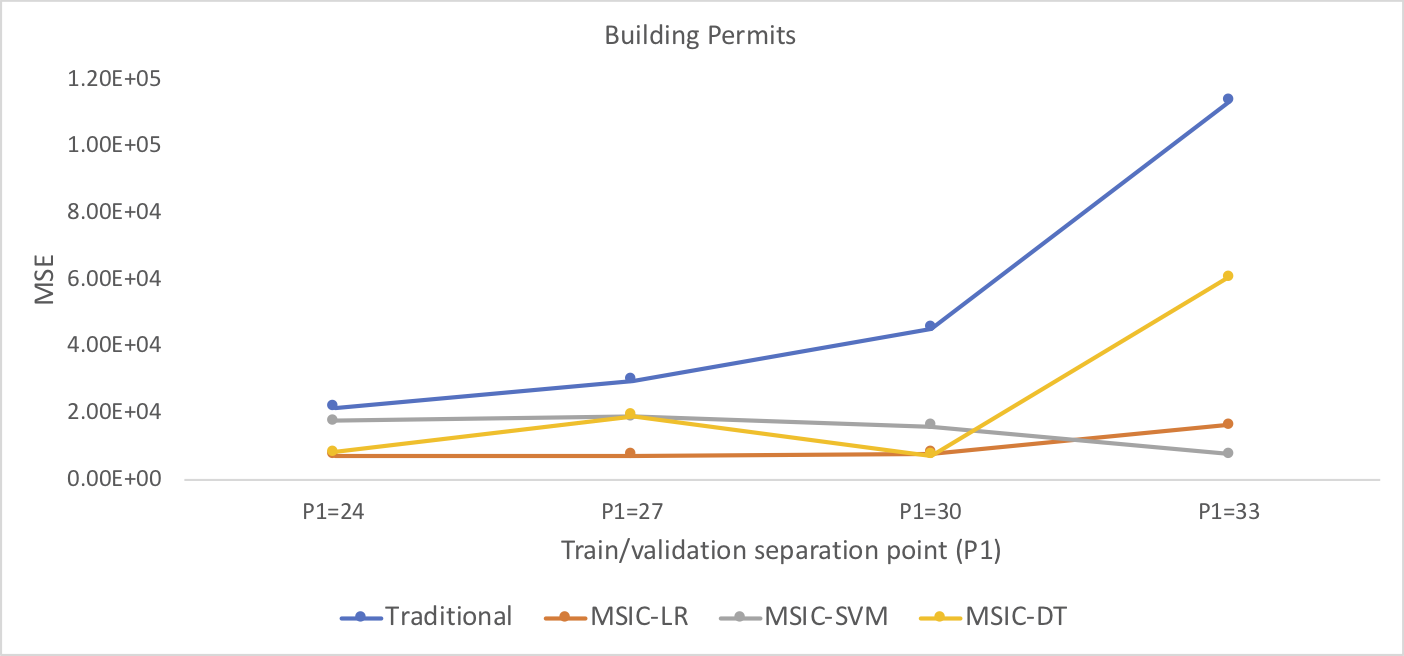}}
        \caption{\label{BP} Performance comparison using "Building Permits" Data}
      \end{figure}
      
\end{landscape}

\begin{figure}[htbp]
        \center{\includegraphics[width=0.8\textwidth]
        {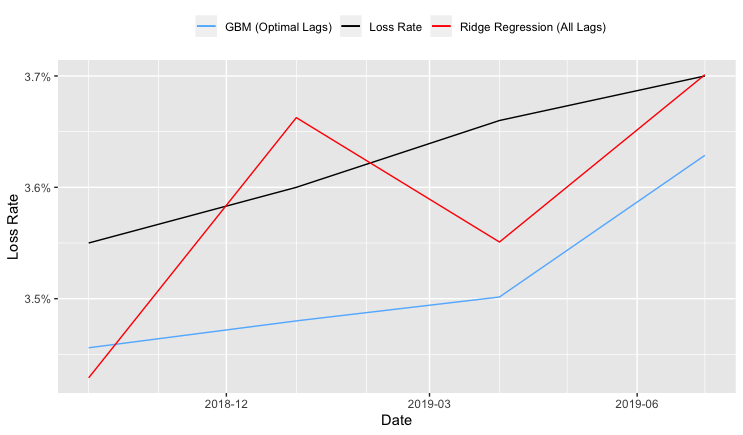}}
        \caption{\label{Preds} Prediction plots for 2018Q2-2019Q2 when using MSIC as forecasting model selection procedure.}
      \end{figure}

Now that the forecasting models are selected for each macroeconomic indicator, and the predictions are made, we use the forecast values as input to the trained models in step 3. Gradient boosting machine was selected as the best performing machine learning model using feature selection with optimal lags, and ridge regression was the winner when using feature selection with all lags. Therefore, these two models are used to generate the final forecasts for the loss rate. MSE results for final forecasts are reported in table \ref{tab:Prediction_MSEs}. Since all the variants of the MISC are generating the same results, we show all the predictions in one figure, which is representative of the results for all the variants of the MSIC algorithm. The prediction plots are shown in figure \ref{Preds}. As the MSE results in table \ref{tab:Prediction_MSEs} show, we achieve significantly low values for MSE using our proposed loss forecasting framework that shows the efficiency of the algorithm. Moreover, looking at figure \ref{Preds}, we see that both variants of our loss forecasting model can closely predict the loss rate values, and it is able to capture the uptrend of the loss rate, which is not possible when using only unemployment rate as the decision variable. Overall, we see that ridge regression with all lags can obtain better results than gradient boosting with optimal lags. The reason is that in the feature selection with all lags, the lags are selected automatically by the model, and we allow the model to use more than one lag from each indicator. This way, more data is available to make predictions. Hence, the Ridge regression with all lags can perform better than gradient boosting with optimal lags. Moreover, Ridge regression is from the family of monotonic and linear machine learning models, which makes it highly interpretable and the assigned coefficients are available for each input variable.

\begin{table}[]
\caption{MSE for predictions resulted from two feature selection approaches (optimal lags and all lags). For each approach, MSE values are reported when using three different variants of MSIC as forecasting model selection procedure}
\label{tab:Prediction_MSEs}
\begin{tabular}{|c|c|c|c|}
\hline
Model & \textbf{MSIC\_LR} & \textbf{MSIC\_SVM} & \textbf{MSIC\_DT} \\ \hline
Gradient Boosting Machine (Optimal Lags)& 1.15E-03 & 1.15E-03 & 1.15E-03  \\ \hline
Ridge Regression (All Lags) & 1.04E-03 & 1.04E-03 & 1.04E-03 \\ \hline
\end{tabular}
\end{table}

Now that we have selected the macroeconomic indicators with significant correlation with charge-off rate and built a prediction model using these values, one may bring up the question that whether the selected macroeconomic indicators are actually the ones causing the fluctuations in the charge-off rate or not. While correlation and causation may exist at the same time, but existence of a correlation does not necessarily implies causation. Causation applies in the situations that an action explicitly triggers another action, but correlation simply implies a relationship. When a correlation exists between two actions, it means that they are related to each other, but it does not necessarily mean that any of them cause the other one. We use the example from the book "Introduction to statistical learning" by \citet{james2013introduction} to explain this issue. Suppose that we are evaluating the correlation between the sales of an ice cream vendor in a beach with number of shark attacks. Interestingly, they have a high correlation, but it does not mean that selling ice cream on the beach causes more shark attacks or vice versa. However, when the weather is hot, people are more attracted to the beaches and consequently, the number of ice cream sales increases. When there are more people on the beach, there is a higher chance of a shark attack and the higher temperature is actually the cause of attracting more people to the beach which results in more shark attacks. This example illustrates the difference between causation and correlation. Our main focus in this research was on correlation rather than causality. The question regarding whether the final significant indicators that were selected to build the model are actually causing the chain of events that leads to the changes in the charge-off rate is left to the experts in the credit card industry and economists. 

\section{Conclusions}
In this paper, we have proposed a machine learning based loss forecasting framework for the credit card industry using macroeconomic indicators. Our goal was to cover macroeconomic indicators from all segments of the economy to make predictions based on a holistic view of the economic conditions. Using the review of the literature and experts' opinion, we selected 19 macroeconomic indicators, which cover consumer, business, and government sections of the economy as input to the proposed loss forecasting framework. The proposed procedure consists of four steps, data preparation, feature selection, model training, and forecasting. We used four machine learning models, namely Lasso regression, Ridge regression, gradient boosting machine, and random forest to develop two versions of the loss forecasting framework. The difference between these two versions is in the utilization of lags from input data. We also applied the proposed model selection procedure in \citet{taghiyeh2020forecasting} (MSIC algorithm) in the forecasting segment of the proposed loss forecasting framework to find the optimal time series forecasting model. To the best of our knowledge, this work is the first that uses an extensive number of the macroeconomic indicators from all segments of the economy to build a machine learning based loss forecasting framework for the US credit card industry. To show the performance of the proposed loss forecasting framework, we used the charge-off data for the top 100 banks in the US ranked by assets from 1985 to 2019, and the data corresponding to selected macroeconomic indicators. We applied the proposed loss forecasting framework on the data, and the final results were very promising. We could achieve the test MSE of 1.15E-03 and 1.04E-03 corresponding to feature selection with optimal lags and feature selection with all lags, respectively, which shows the effectiveness of the proposed algorithm in forecasting the loss rate. The final fit for the prediction period shows that we could closely predict the actual values of the loss rate and the uptrend of the loss rate could be captured by our proposed model, which was not possible in the conventional version of the credit card loss forecasting frameworks that only use the unemployment rate as the decision variable.

In the future, we aim to further improve the proposed loss forecasting model in this paper by adding more machine learning models, such as deep neural networks, long-short term memory (LSTM) model, and extreme gradient boosting to the benchmark models and see if we can make more accurate forecasts. Additionally, more feature selection procedures can be explored to improve the feature selection step of the loss forecasting framework. The other future line of research would be to perform a more exhaustive number of transformation for the macroeconomic indicators to see if a better data transformation can be found to improve the efficiency of the algorithm further, as the final results are sensitive to these transformations. Another interesting future research path is to analyze the credit card charge-off rates due to the rapid changes in the economy caused by the Coronavirus pandemic and adjust the model accordingly.


\bibliography{Loss_Forecasting}

\begin{thebibliography}{38}
\expandafter\ifx\csname natexlab\endcsname\relax\def\natexlab#1{#1}\fi
\providecommand{\url}[1]{\texttt{#1}}
\providecommand{\href}[2]{#2}
\providecommand{\path}[1]{#1}
\providecommand{\DOIprefix}{doi:}
\providecommand{\ArXivprefix}{arXiv:}
\providecommand{\URLprefix}{URL: }
\providecommand{\Pubmedprefix}{pmid:}
\providecommand{\doi}[1]{\href{http://dx.doi.org/#1}{\path{#1}}}
\providecommand{\Pubmed}[1]{\href{pmid:#1}{\path{#1}}}
\providecommand{\bibinfo}[2]{#2}
\ifx\xfnm\relax \def\xfnm[#1]{\unskip,\space#1}\fi
\bibitem[{Agarwal and Liu(2003)}]{agarwal2003determinants}
\bibinfo{author}{Agarwal, S.}, \bibinfo{author}{Liu, C.}, \bibinfo{year}{2003}.
\newblock \bibinfo{title}{Determinants of credit card delinquency and
  bankruptcy: Macroeconomic factors}.
\newblock \bibinfo{journal}{Journal of Economics and Finance}
  \bibinfo{volume}{27}, \bibinfo{pages}{75--84}.
\bibitem[{Ausubel(1997)}]{ausubel1997credit}
\bibinfo{author}{Ausubel, L.M.}, \bibinfo{year}{1997}.
\newblock \bibinfo{title}{Credit card defaults, credit card profits, and
  bankruptcy}.
\newblock \bibinfo{journal}{Am. Bankr. LJ} \bibinfo{volume}{71},
  \bibinfo{pages}{249}.
\bibitem[{Bellotti and Crook(2012)}]{bellotti2012loss}
\bibinfo{author}{Bellotti, T.}, \bibinfo{author}{Crook, J.},
  \bibinfo{year}{2012}.
\newblock \bibinfo{title}{Loss given default models incorporating macroeconomic
  variables for credit cards}.
\newblock \bibinfo{journal}{International Journal of Forecasting}
  \bibinfo{volume}{28}, \bibinfo{pages}{171--182}.
\bibitem[{Bellotti and Crook(2013)}]{bellotti2013forecasting}
\bibinfo{author}{Bellotti, T.}, \bibinfo{author}{Crook, J.},
  \bibinfo{year}{2013}.
\newblock \bibinfo{title}{Forecasting and stress testing credit card default
  using dynamic models}.
\newblock \bibinfo{journal}{International Journal of Forecasting}
  \bibinfo{volume}{29}, \bibinfo{pages}{563--574}.
\bibitem[{Berge and Jord{\`a}(2011)}]{berge2011evaluating}
\bibinfo{author}{Berge, T.J.}, \bibinfo{author}{Jord{\`a}, {\`O}.},
  \bibinfo{year}{2011}.
\newblock \bibinfo{title}{Evaluating the classification of economic activity
  into recessions and expansions}.
\newblock \bibinfo{journal}{American Economic Journal: Macroeconomics}
  \bibinfo{volume}{3}, \bibinfo{pages}{246--77}.
\bibitem[{{Board of Governors of the Federal Reserve System
  (U.S.)}(2020)}]{CO:2020}
\bibinfo{author}{{Board of Governors of the Federal Reserve System (U.S.)}},
  \bibinfo{year}{2020}.
\newblock \bibinfo{title}{Charge-off rate on credit card loans, top 100 banks
  ranked by assets [corcct100s]}.
\newblock \bibinfo{note}{Retrieved from FRED, Federal Reserve Bank of St.
  Louis; \url{https://fred.stlouisfed.org/series/CORCCT100S}}.
\bibitem[{Censky(2010)}]{censky2010consumer}
\bibinfo{author}{Censky, A.}, \bibinfo{year}{2010}.
\newblock \bibinfo{title}{Consumer confidence slumps in september}.
\newblock \bibinfo{journal}{CNN Money} .
\bibitem[{Dalio(2018)}]{RayDalio2018}
\bibinfo{author}{Dalio, R.}, \bibinfo{year}{2018}.
\newblock \bibinfo{title}{Principles for navigating big debt crises}.
\newblock \bibinfo{publisher}{Westport, CT : Bridgewater}.
\bibitem[{Debbaut et~al.(2016)Debbaut, Ghent and Kudlyak}]{debbaut2016card}
\bibinfo{author}{Debbaut, P.}, \bibinfo{author}{Ghent, A.},
  \bibinfo{author}{Kudlyak, M.}, \bibinfo{year}{2016}.
\newblock \bibinfo{title}{The card act and young borrowers: The effects and the
  affected}.
\newblock \bibinfo{journal}{Journal of Money, Credit and Banking}
  \bibinfo{volume}{48}, \bibinfo{pages}{1495--1513}.
\bibitem[{Desai et~al.(2014)Desai, Elliehausen and Lawrence}]{desai2014county}
\bibinfo{author}{Desai, C.A.}, \bibinfo{author}{Elliehausen, G.},
  \bibinfo{author}{Lawrence, E.C.}, \bibinfo{year}{2014}.
\newblock \bibinfo{title}{On the county-level credit outcome beta}.
\newblock \bibinfo{journal}{Journal of Financial Services Research}
  \bibinfo{volume}{45}, \bibinfo{pages}{201--218}.
\bibitem[{Emekter et~al.(2015)Emekter, Tu, Jirasakuldech and
  Lu}]{emekter2015evaluating}
\bibinfo{author}{Emekter, R.}, \bibinfo{author}{Tu, Y.},
  \bibinfo{author}{Jirasakuldech, B.}, \bibinfo{author}{Lu, M.},
  \bibinfo{year}{2015}.
\newblock \bibinfo{title}{Evaluating credit risk and loan performance in online
  peer-to-peer (p2p) lending}.
\newblock \bibinfo{journal}{Applied Economics} \bibinfo{volume}{47},
  \bibinfo{pages}{54--70}.
\bibitem[{Estrella and Hardouvelis(1991)}]{estrella1991term}
\bibinfo{author}{Estrella, A.}, \bibinfo{author}{Hardouvelis, G.A.},
  \bibinfo{year}{1991}.
\newblock \bibinfo{title}{The term structure as a predictor of real economic
  activity}.
\newblock \bibinfo{journal}{The journal of Finance} \bibinfo{volume}{46},
  \bibinfo{pages}{555--576}.
\bibitem[{Estrella and Mishkin(1998)}]{estrella1998predicting}
\bibinfo{author}{Estrella, A.}, \bibinfo{author}{Mishkin, F.S.},
  \bibinfo{year}{1998}.
\newblock \bibinfo{title}{Predicting us recessions: Financial variables as
  leading indicators}.
\newblock \bibinfo{journal}{Review of Economics and Statistics}
  \bibinfo{volume}{80}, \bibinfo{pages}{45--61}.
\bibitem[{Evans and Schmalensee(2005)}]{evans2005paying}
\bibinfo{author}{Evans, D.S.}, \bibinfo{author}{Schmalensee, R.},
  \bibinfo{year}{2005}.
\newblock \bibinfo{title}{Paying with plastic: the digital revolution in buying
  and borrowing}.
\newblock \bibinfo{publisher}{Mit Press}.
\bibitem[{Figlewski et~al.(2012)Figlewski, Frydman and
  Liang}]{figlewski2012modeling}
\bibinfo{author}{Figlewski, S.}, \bibinfo{author}{Frydman, H.},
  \bibinfo{author}{Liang, W.}, \bibinfo{year}{2012}.
\newblock \bibinfo{title}{Modeling the effect of macroeconomic factors on
  corporate default and credit rating transitions}.
\newblock \bibinfo{journal}{International Review of Economics \& Finance}
  \bibinfo{volume}{21}, \bibinfo{pages}{87--105}.
\bibitem[{Fung and Wong(2002)}]{fung2002modeling}
\bibinfo{author}{Fung, T.}, \bibinfo{author}{Wong, M.}, \bibinfo{year}{2002}.
\newblock \bibinfo{title}{Modeling credit card charge-off ratios: The case of
  hong kong}.
\newblock \bibinfo{journal}{City University of Hong Kong: Department of
  Economics \& Finance} .
\bibitem[{Giesecke et~al.(2011)Giesecke, Longstaff, Schaefer and
  Strebulaev}]{giesecke2011corporate}
\bibinfo{author}{Giesecke, K.}, \bibinfo{author}{Longstaff, F.A.},
  \bibinfo{author}{Schaefer, S.}, \bibinfo{author}{Strebulaev, I.},
  \bibinfo{year}{2011}.
\newblock \bibinfo{title}{Corporate bond default risk: A 150-year perspective}.
\newblock \bibinfo{journal}{Journal of Financial Economics}
  \bibinfo{volume}{102}, \bibinfo{pages}{233--250}.
\bibitem[{Gross and Souleles(2002)}]{gross2002empirical}
\bibinfo{author}{Gross, D.B.}, \bibinfo{author}{Souleles, N.S.},
  \bibinfo{year}{2002}.
\newblock \bibinfo{title}{An empirical analysis of personal bankruptcy and
  delinquency}.
\newblock \bibinfo{journal}{The Review of Financial Studies}
  \bibinfo{volume}{15}, \bibinfo{pages}{319--347}.
\bibitem[{Guo et~al.(2016)Guo, Zhou, Luo, Liu and Xiong}]{guo2016instance}
\bibinfo{author}{Guo, Y.}, \bibinfo{author}{Zhou, W.}, \bibinfo{author}{Luo,
  C.}, \bibinfo{author}{Liu, C.}, \bibinfo{author}{Xiong, H.},
  \bibinfo{year}{2016}.
\newblock \bibinfo{title}{Instance-based credit risk assessment for investment
  decisions in p2p lending}.
\newblock \bibinfo{journal}{European Journal of Operational Research}
  \bibinfo{volume}{249}, \bibinfo{pages}{417--426}.
\bibitem[{Guseva and Rona-Tas(2001)}]{guseva2001uncertainty}
\bibinfo{author}{Guseva, A.}, \bibinfo{author}{Rona-Tas, A.},
  \bibinfo{year}{2001}.
\newblock \bibinfo{title}{Uncertainty, risk, and trust: Russian and american
  credit card markets compared}.
\newblock \bibinfo{journal}{American sociological review} ,
  \bibinfo{pages}{623--646}.
\bibitem[{James et~al.(2013)James, Witten, Hastie and
  Tibshirani}]{james2013introduction}
\bibinfo{author}{James, G.}, \bibinfo{author}{Witten, D.},
  \bibinfo{author}{Hastie, T.}, \bibinfo{author}{Tibshirani, R.},
  \bibinfo{year}{2013}.
\newblock \bibinfo{title}{An introduction to statistical learning}. volume
  \bibinfo{volume}{112}.
\newblock \bibinfo{publisher}{Springer}.
\bibitem[{Kim et~al.(2017)Kim, Won and Kim}]{kim2017additional}
\bibinfo{author}{Kim, J.W.}, \bibinfo{author}{Won, S.}, \bibinfo{author}{Kim,
  J.I.}, \bibinfo{year}{2017}.
\newblock \bibinfo{title}{Additional credit for liquidity-constrained
  individuals: High-interest consumer credit in korea}.
\newblock \bibinfo{journal}{Emerging Markets Finance and Trade}
  \bibinfo{volume}{53}, \bibinfo{pages}{109--127}.
\bibitem[{Kovrijnykh and Livshits(2017)}]{kovrijnykh2017screening}
\bibinfo{author}{Kovrijnykh, N.}, \bibinfo{author}{Livshits, I.},
  \bibinfo{year}{2017}.
\newblock \bibinfo{title}{Screening as a unified theory of delinquency,
  renegotiation, and bankruptcy}.
\newblock \bibinfo{journal}{International Economic Review}
  \bibinfo{volume}{58}, \bibinfo{pages}{499--527}.
\bibitem[{Leow and Crook(2014)}]{leow2014intensity}
\bibinfo{author}{Leow, M.}, \bibinfo{author}{Crook, J.}, \bibinfo{year}{2014}.
\newblock \bibinfo{title}{Intensity models and transition probabilities for
  credit card loan delinquencies}.
\newblock \bibinfo{journal}{European Journal of Operational Research}
  \bibinfo{volume}{236}, \bibinfo{pages}{685--694}.
\bibitem[{Levanon et~al.(2011)Levanon, Ozyildirim, Manini, Schaitkin and
  Tanchua}]{levanon2011using}
\bibinfo{author}{Levanon, G.}, \bibinfo{author}{Ozyildirim, A.},
  \bibinfo{author}{Manini, J.C.}, \bibinfo{author}{Schaitkin, B.},
  \bibinfo{author}{Tanchua, J.}, \bibinfo{year}{2011}.
\newblock \bibinfo{title}{Using a leading credit index to predict turning
  points in the us business cycle}, in: \bibinfo{booktitle}{The Conference
  Board Economics Program Working Paper}.
\bibitem[{Liu and Xu(2003)}]{liu2003predictive}
\bibinfo{author}{Liu, J.}, \bibinfo{author}{Xu, X.E.}, \bibinfo{year}{2003}.
\newblock \bibinfo{title}{The predictive power of economic indicators in
  consumer credit risk management}.
\newblock \bibinfo{journal}{Rma Journal} \bibinfo{volume}{86},
  \bibinfo{pages}{40--45}.
\bibitem[{Malekipirbazari and Aksakalli(2015)}]{malekipirbazari2015risk}
\bibinfo{author}{Malekipirbazari, M.}, \bibinfo{author}{Aksakalli, V.},
  \bibinfo{year}{2015}.
\newblock \bibinfo{title}{Risk assessment in social lending via random
  forests}.
\newblock \bibinfo{journal}{Expert Systems with Applications}
  \bibinfo{volume}{42}, \bibinfo{pages}{4621--4631}.
\bibitem[{Maziba{\c{s}} and Tuna(2017)}]{mazibacs2017understanding}
\bibinfo{author}{Maziba{\c{s}}, M.}, \bibinfo{author}{Tuna, Y.},
  \bibinfo{year}{2017}.
\newblock \bibinfo{title}{Understanding the recent growth in consumer loans and
  credit cards in emerging markets: Evidence from turkey}.
\newblock \bibinfo{journal}{Emerging Markets Finance and Trade}
  \bibinfo{volume}{53}, \bibinfo{pages}{2333--2346}.
\bibitem[{Mian and Sufi(2011)}]{mian2011house}
\bibinfo{author}{Mian, A.}, \bibinfo{author}{Sufi, A.}, \bibinfo{year}{2011}.
\newblock \bibinfo{title}{House prices, home equity-based borrowing, and the us
  household leverage crisis}.
\newblock \bibinfo{journal}{American Economic Review} \bibinfo{volume}{101},
  \bibinfo{pages}{2132--56}.
\bibitem[{Musto and Souleles(2006)}]{musto2006portfolio}
\bibinfo{author}{Musto, D.K.}, \bibinfo{author}{Souleles, N.S.},
  \bibinfo{year}{2006}.
\newblock \bibinfo{title}{A portfolio view of consumer credit}.
\newblock \bibinfo{journal}{Journal of Monetary Economics}
  \bibinfo{volume}{53}, \bibinfo{pages}{59--84}.
\bibitem[{Ng(2014)}]{ng2014boosting}
\bibinfo{author}{Ng, S.}, \bibinfo{year}{2014}.
\newblock \bibinfo{title}{Boosting recessions}.
\newblock \bibinfo{journal}{Canadian Journal of Economics/Revue canadienne
  d'{\'e}conomique} \bibinfo{volume}{47}, \bibinfo{pages}{1--34}.
\bibitem[{Peterson(2017)}]{peterson2017bestnormalize}
\bibinfo{author}{Peterson, R.}, \bibinfo{year}{2017}.
\newblock \bibinfo{title}{bestnormalize: A suite of normalizing
  transformations}.
\newblock \bibinfo{journal}{R package version} \bibinfo{volume}{3}.
\bibitem[{Rubaszek and Serwa(2014)}]{rubaszek2014determinants}
\bibinfo{author}{Rubaszek, M.}, \bibinfo{author}{Serwa, D.},
  \bibinfo{year}{2014}.
\newblock \bibinfo{title}{Determinants of credit to households: An approach
  using the life-cycle model}.
\newblock \bibinfo{journal}{Economic Systems} \bibinfo{volume}{38},
  \bibinfo{pages}{572--587}.
\bibitem[{Schmitt(2000)}]{schmitt2000does}
\bibinfo{author}{Schmitt, E.D.}, \bibinfo{year}{2000}.
\newblock \bibinfo{title}{Does rising consumer debt signal future recessions?:
  Testing the causal relationship between consumer debt and the economy}.
\newblock \bibinfo{journal}{Atlantic Economic Journal} \bibinfo{volume}{28},
  \bibinfo{pages}{333--345}.
\bibitem[{Stavins et~al.(2000)}]{stavins2000credit}
\bibinfo{author}{Stavins, J.}, et~al., \bibinfo{year}{2000}.
\newblock \bibinfo{title}{Credit card borrowing, delinquency, and personal
  bankruptcy}.
\newblock \bibinfo{journal}{New England Economic Review} ,
  \bibinfo{pages}{15--30}.
\bibitem[{Stock and Watson(2002)}]{stock2002forecasting}
\bibinfo{author}{Stock, J.H.}, \bibinfo{author}{Watson, M.W.},
  \bibinfo{year}{2002}.
\newblock \bibinfo{title}{Forecasting using principal components from a large
  number of predictors}.
\newblock \bibinfo{journal}{Journal of the American statistical association}
  \bibinfo{volume}{97}, \bibinfo{pages}{1167--1179}.
\bibitem[{Taghiyeh et~al.(2020)Taghiyeh, Lengacher and
  Handfield}]{taghiyeh2020forecasting}
\bibinfo{author}{Taghiyeh, S.}, \bibinfo{author}{Lengacher, D.C.},
  \bibinfo{author}{Handfield, R.B.}, \bibinfo{year}{2020}.
\newblock \bibinfo{title}{Forecasting model selection using intermediate
  classification: Application to monarchfx corporation}.
\newblock \bibinfo{journal}{Expert Systems with Applications} ,
  \bibinfo{pages}{113371}.
\bibitem[{Taghiyeh and Xu(2016)}]{taghiyeh2016new}
\bibinfo{author}{Taghiyeh, S.}, \bibinfo{author}{Xu, J.}, \bibinfo{year}{2016}.
\newblock \bibinfo{title}{A new particle swarm optimization algorithm for noisy
  optimization problems}.
\newblock \bibinfo{journal}{Swarm Intelligence} \bibinfo{volume}{10},
  \bibinfo{pages}{161--192}.

\end{thebibliography}

\end{document}